\documentclass[11pt]{article}

\usepackage[final]{acl}

\usepackage{times}
\usepackage{latexsym}
\usepackage[T1]{fontenc}
\usepackage[utf8]{inputenc}
\usepackage{microtype}
\usepackage{inconsolata}
\usepackage{graphicx}
\usepackage{algorithm}
\usepackage{algpseudocode}

\usepackage{booktabs}      
\usepackage{amsmath}
\usepackage{amssymb}
\usepackage{multirow}
\usepackage{xcolor}
\usepackage{url}
\usepackage{subcaption}
\usepackage{listings}

\lstdefinestyle{promptstyle}{
  basicstyle=\footnotesize\ttfamily,
  breaklines=true,
  breakindent=0pt,
  columns=fullflexible,
  keepspaces=true,
  frame=none,
  aboveskip=8pt, belowskip=8pt,
  xleftmargin=12pt, xrightmargin=3pt,
  literate=%
    {“}{{``}}1 {”}{{''}}1
    {‘}{{`}}1 {’}{{'}}1
    {–}{{--}}1 {—}{{---}}1
}
\lstnewenvironment{promptbox}{\lstset{style=promptstyle}}{}

\graphicspath{{figures/}}

\title{The Patent Lottery: Generating a Structured Judgment Space from Single-Sample Examination}

\author{Yehoon Jang$^{1}$, Irvan Zidny$^{1}$,
        Sungbin Lim$^{2}$,
        Hyun-seok Min$^{3}$\footnotemark[1],
        Sungchul Choi$^{1}$\thanks{Corresponding authors} \\
  $^{1}$Pukyong National University \quad
   $^{2}$Korea University \& NST \quad
  $^{3}$Tomocube  \\
  \texttt{\{jangyh0420, zidny4399\}@pukyong.ac.kr} \quad
  \texttt{sungbin@korea.ac.kr} \\
  \texttt{min6284@gmail.com} \quad
  \texttt{sc82.choi@pknu.ac.kr}}

\begin{document}
\maketitle

\begin{abstract}
Patent examination outcomes hinge on which examiner a case lands with---a ``patent lottery'' with real economic stakes---yet the real world only ever produces one examiner, one decision, no counterfactual. We construct computational examiner populations that generate the \emph{judgment space} for each patent case: the full distribution of professional positions the case could elicit. Using 14 AI examiners varying along five dimensions in an orthogonal design, we examine 100 PTAB cases (1,400 decisions). The judgment space shows structured dimension effects: examination strictness dominates, with strict profiles overriding other characteristics, while lenient outcomes require multiple lenient dimensions to align. It contains the real institutional outcome in 99 of 100 cases, and within-space disagreements trace to specific examiner dimensions rather than random noise. Trace-level analysis locates \S103 as the stage where disagreement both concentrates and translates directly into divergent final decisions. 
\end{abstract}

\section{Introduction}
\label{sec:intro}

When an inventor files an application at the United States Patent and Trademark Office (USPTO), the application is assigned to an examiner through a process that is effectively random within technology areas, and that assignment has measurable consequences.

Examiner-level variation has been documented along multiple dimensions. Individual grant rates differ widely \citep{lemley2012examiner, frakes2015patent}, with downstream consequences for patent scope and market reactions \citep{schuster2023sweepstakes}, litigation rates \citep{feng2020crafting}, and innovation quality \citep{dyer2024effect}. The USPTO grants too many invalid patents, with examiner-level tendencies as a key driver \citep{frakes2015patent}.

The economic magnitude of this variation is substantial. The USPTO examines over 600,000 applications a year, so even modest examiner inconsistency moves thousands of decisions on which venture funding, R\&D investment, and litigation strategy turn. A recent GAO report confirms persistent quality challenges in USPTO examination \citep{gao2025intellectual}.

Every empirical finding above rests on single-sample observations: one examiner, one application, one outcome. The counterfactual, what a different examiner would have decided on the same application, is never observed. So while the patent lottery is well established, why examiners diverge and which characteristics drive that divergence remain open, because the controlled experiment of assigning one patent to many examiners at once has been infeasible.

Computational examiner populations offer a way around the single-sample constraint. Instead of observing one examiner's decision and inferring the rest, we can have the same patent examined by a population of examiners, each with different but empirically-grounded characteristics, and generate the full distribution of outcomes. We call this distribution the \textbf{judgment space}: the set of professional positions a single case elicits across varied examiner profiles, each reached by applying a legal framework to the case evidence through explicit reasoning.

Three properties must hold for this space to be meaningful: \textbf{coverage} (real outcomes fall inside), \textbf{profile-attributability} (disagreements trace to identifiable dimensions), and \textbf{legal coherence} (each position reasons under the appropriate statute, ties its conclusions to evidence, and stays consistent across stages).

\begin{figure*}[t]
\centering
\includegraphics[width=\textwidth]{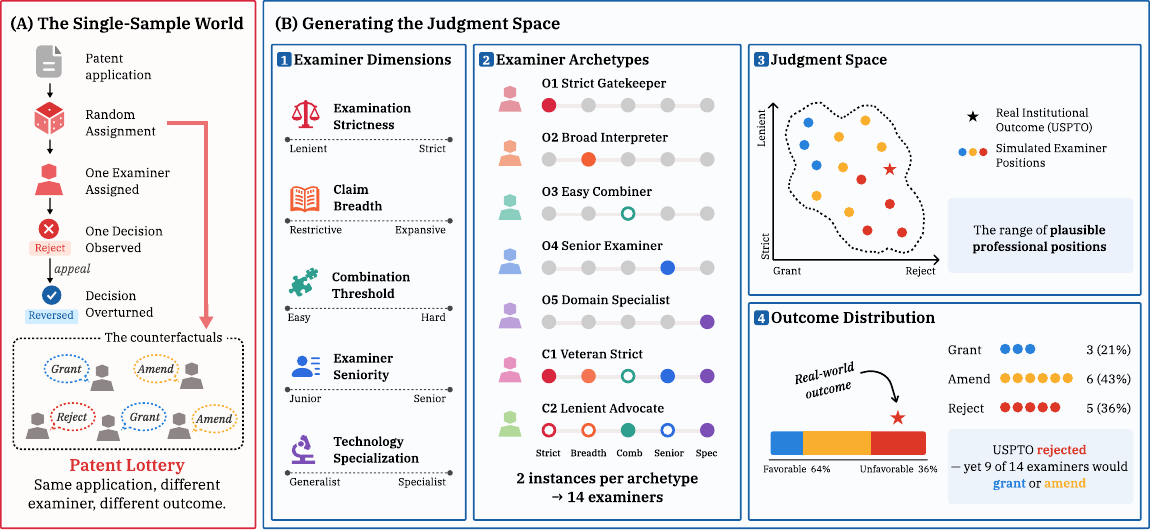}
\caption{From single-sample examination to structured judgment space. \textbf{(A) The single-sample world:} a patent application is randomly assigned to one USPTO examiner, and only that examiner's decision is observed; the outcomes other examiners would have reached stay unobservable, even after PTAB appeal. \textbf{(B) Generating the judgment space:} we instead examine the same case with a population of 14 examiners (7 archetypes, 2 instances each) defined by five skill parameters in an orthogonal-plus-composite design (O1--O5 single-dimension, C1--C2 composite). Filled markers in panel 2 mark values above the 0.50 baseline and hollow markers values below it. All examiners share an 8-stage pipeline and differ in their parameter values, producing a distribution of professional positions, the judgment space, in which the real institutional outcome ($\star$) sits among the simulated ones.}
\label{fig:overview}
\end{figure*}

We present the first systematic study of judgment space generation for patent examination, using 100 PTAB cases examined by 14 AI examiners (1,400 decisions). The examiner population follows an orthogonal-plus-composite design. Five single-dimension archetypes isolate the causal effect of each characteristic, and two composite archetypes capture realistic multi-trait interaction. Examiners differ in quantitative thresholds within shared legal frameworks, so diversity reflects differences in professional judgment rather than persona-label variation typical of prior LLM simulation work (\S\ref{sec:related}).

\paragraph{Contributions.} The contribution of this paper is methodological. We propose a task and a framework for generating examiner judgment spaces, and we apply them to real PTAB cases to establish properties of examiner disagreement that single-sample data cannot reveal.

\textbf{Task formulation.} We move from single-decision prediction to judgment space generation. The input is one application, the output is the set of decisions, profiles, and traces produced by a population of differently calibrated examiners, and evaluation rests on three properties of that set, namely coverage, profile-attributability, and legal coherence, which remain testable when the data are single-sample.

\textbf{Skill-parameter framework.} An examiner is a vector of five numerical parameters, each grounded in empirical examiner research and attached to a specific pipeline stage, rather than a natural-language persona label. The orthogonal-plus-composite population design lets outcome differences be attributed to individual dimensions.

\textbf{Counterfactual evidence on the patent lottery.} Because every examiner profile sees the same case, the experiment observes what single-sample data never can, namely how one application fares under other examiners. On 100 PTAB cases and 1{,}400 decisions, four results follow from the experiment rather than from the design. Strictness moves the favorable-decision rate 5 to 24 times more than any other dimension. Lenient dimensions compound super-additively while strict ones override. Disagreement concentrates in the \S103 obviousness stage and carries through to the final decision. The ordering of the archetypes at the lenient and strict ends is preserved across six LLM backbones of different sizes and families. In the simulated population, the lottery therefore has a definite shape. One dimension dominates, it acts asymmetrically, and it enters at one stage. The space also contains the real PTAB outcome in 99 of 100 cases, and an institutional-oversight intervention reshapes it unevenly across archetypes (\S\ref{sec:oversight}).

\section{Related Work}
\label{sec:related}

\subsection{Empirical Studies of Examiner Variation}

A substantial body of empirical research has documented examiner-level variation in patent outcomes. \citet{lemley2012examiner} report wide individual grant-rate variation and that more experienced examiners grant at higher rates and cite less prior art. \citet{frakes2017time} show that examiner promotion reduces per-case time allocation and increases grant rates, with the additional patents being of below-average quality. At the system level, \citet{frakes2015patent} provide quasi-experimental evidence of systematic overgranting driven in part by examiner-level tendencies. Separately, \citet{righi2019examiner} find that more specialized examiners have lower grant rates, a separate channel of examination variation.

These studies share a methodological constraint. Specifically, they observe single-sample outcomes, with one examiner's decision per application, and infer population-level patterns through statistical aggregation. The counterfactual question (``what would a different examiner have decided on this same application?'') cannot be directly answered. To address this, our work complements this literature by generating the full judgment space for each case.

\subsection{LLM for Patent Reasoning}

Recent NLP work on patent reasoning has primarily framed the task as single-decision prediction. The Harvard USPTO Patent Dataset \citep{suzgun2023harvard} introduced patent acceptance prediction, and subsequent work has extended this single-decision framing to PTAB issue type classification \citep{jang2025pilot}, \S112(b) definiteness assessment \citep{knappich2025pedantic}, and reconstruction of step-by-step examination trails \citep{lim2025panorama}.

Like the empirical studies above, these approaches share a structural limitation. Each model produces one prediction per application, mirroring the real-world constraint of one examiner per case and providing no access to the counterfactual. To our knowledge, no prior work has generated the \textit{judgment space} for patent examination.

\subsection{LLM-Based Generation of Professional Judgment}
\label{sec:related-llmgen}

Beyond single-sample prediction, a separate line of work has explored population-level simulation of professional judgment, covering peer review \citep{jin2024agentreview}, legal proceedings \citep{chen2025agentcourt,wang2025law}, clinical interaction \citep{kyung2025patientsim}, and economic behavior \citep{horton2024homo,bui2025mop}. These approaches typically model diversity through persona descriptions, using surface-level trait labels such as ``strict examiner'' or ``lenient reviewer''. However, this approach has limitations.

Persona effects are \textit{weak}: \citet{hucollier2024quantifying} find that persona variables explain only a small fraction of LLM annotation variance, suggesting limited control over model behavior. Beyond weakness, labels \textit{collapse}: agents with different labels often converge on similar outputs, a phenomenon termed ``diversity collapse'' \citep{wang2026parametric}. And finally, they are \textit{semantically unstable}: the same label (e.g., ``strict'') can be interpreted differently across models and contexts, yielding uncontrollable variation rather than structured diversity \citep{lin2025six,agnew2024illusion}. Our baseline experiments provide supporting evidence for each failure mode (Appendix~\ref{app:baseline}).

We address these limitations as follows. We model diversity through \textit{skill parameters}, quantitative thresholds within shared legal frameworks, rather than personality labels. An orthogonal design structurally guarantees that diversity persists across runs. And we frame the goal as judgment space generation rather than human substitution: the output is a distribution of professional positions, not a prediction of any single outcome. This skill-parameter approach is also consistent with cognitive science findings on expert decision-making \citep{kahneman2009conditions}, which links expert disagreement to differences in application thresholds rather than personality traits. A full comparison table and additional discussion are in Appendix~\ref{app:related}.

\section{Method}
\label{sec:method}

\subsection{Background}

U.S. patent examination evaluates applications against four statutory requirements, namely \S102 (novelty), \S103 (non-obviousness), \S101 (eligibility), and \S112 (written description), following the Manual of Patent Examining Procedure (MPEP). Of these, \S103 involves the most examiner discretion: it requires judging whether a combination of prior art would have been obvious, a standard the Supreme Court deliberately left flexible in \textit{KSR v.~Teleflex} (2007). Rejected applicants may appeal to PTAB, whose decisions provide our ground truth. Because PTAB reviews only appealed cases, this ground truth, and our coverage claim, applies to the appealed subpopulation rather than to examination as a whole. Full legal background, including MPEP section correspondences and KSR rationale expansion, is in Appendix~\ref{app:pipeline}.

\subsection{Design Principles}
\label{sec:design}

Our framework targets \textit{judgment space generation} rather than predicting any single decision: each real application receives one observed outcome, but many differently calibrated examiners would have produced different decisions.

\textit{Task.} A case $c$ consists of the claims, the specification, and the cited prior-art references. The PTAB outcome $y(c)$ is not part of the input; we use it only to evaluate the generated space. An examiner is a profile $\theta \in [0,1]^5$ over the five dimensions in \S\ref{sec:examiner-dimensions}. All examiners run the same eight-stage pipeline $F$ (Appendix~\ref{app:pipeline}). It returns a decision $d \in \{\text{grant}, \text{amend}, \text{reject}\}$ and a trace $T$ that holds the structured output of every stage. The population $\Theta = \{\theta_1, \ldots, \theta_{14}\}$ (\S\ref{sec:population}) is generated once and reused for every case. The judgment space of $c$ is then
\begin{equation}
J(c) = \{(d_i, \theta_i, T_i)\}_{i=1}^{14}, \qquad (d_i, T_i) = F(c, \theta_i).
\label{eq:jspace}
\end{equation}
We do not collapse the 14 tuples into a single prediction.

\textit{Validation.} We evaluate $J(c)$ on three properties: (i) \textit{coverage}, whether the direction of $y(c)$ appears among the $d_i$; (ii) \textit{profile-attributability}, whether differences among the $d_i$ can be traced to the $\theta_i$ rather than to stochastic noise (\S\ref{sec:coverage}); and (iii) \textit{legal coherence}, whether each $T_i$ cites the right statute, ties its conclusions to evidence, and stays consistent from one stage to the next (Appendix~\ref{app:cascade}). The framework complements empirical PTAB research; it does not substitute for it.

\textit{Design versus measurement.} The design fixes the five dimensions, the stage at which each one enters the pipeline, and the sign of two effects. A strict examiner is instructed to resolve uncertain evidence toward rejection, and a high combination threshold requires explicit teaching before references can be combined (Appendix~\ref{app:pipeline}). Breadth, seniority, and specialization enter only as descriptive labels with no stated direction. The negative sign of the strictness effect, and the fact that disagreement follows strictness, are therefore checks that the backbone obeys its parameters rather than findings about examination. Everything else is measured. This includes the size of each effect and the ranking across dimensions, the direction of the three unsigned dimensions, how dimensions combine in composite profiles (\S\ref{sec:dimension-effects}), where in the pipeline disagreement concentrates (\S\ref{sec:process}), whether the PTAB outcome is covered (\S\ref{sec:coverage}), and whether the pattern holds across backbones (\S\ref{sec:robustness}).

\subsection{Examiner Dimensions}
\label{sec:examiner-dimensions}

Each examiner is parameterized along five dimensions: examination strictness, claim interpretation breadth, obviousness combination threshold, examiner seniority, and technology specialization. We select these on a strict criterion: each must have been identified as a statistically significant predictor of examination outcomes in peer-reviewed empirical research. Full empirical basis, legal grounding, and pipeline-stage mapping for each dimension are in Appendix~\ref{app:pipeline}.

\textit{How each dimension enters the pipeline.} Every dimension value is printed, with a short label, in the persona block that precedes each stage. Two dimensions additionally carry a rule. Combination threshold modulates the \S103 obviousness analysis, the stage we predict to be the primary locus of examiner discretion. A higher threshold makes a \S103 rejection harder to sustain, since more evidence is required before prior art can be combined, and this pushes outcomes toward leniency. This is why the lenient advocate C2 has a high combination threshold of 0.70 even though high values on the other dimensions indicate strictness. Strictness governs the final decision, where uncertain grounds are resolved, and its value is also passed to the \S102 stage. The other three dimensions carry no rule. Claim breadth enters the claim interpretation stage as a numeric parameter that scales the scope of reading. Tech specialization is set numerically at the \S112 stage. Seniority is carried as a numeric value in the persona block. How the backbone acts on these three values is measured, not designed (\S\ref{sec:design}).

\subsection{Examiner Population}
\label{sec:population}

\begin{table}[h]
\centering
\caption{Orthogonal archetype parameters. O1: strict gatekeeper, O2: broad interpreter, O3: easy combiner, O4: senior examiner, O5: domain specialist. Bold values indicate the isolated extreme dimension.}
\label{tab:archetypes}
\small
\begin{tabular}{@{}cccccc@{}}
\toprule
\textbf{ID} & \textbf{Strict.} & \textbf{Breadth} & \textbf{Comb.} & \textbf{Senior.} & \textbf{Spec.} \\
\midrule
O1 & \textbf{0.85} & 0.50 & 0.50 & 0.50 & 0.50 \\
O2 & 0.50 & \textbf{0.80} & 0.50 & 0.50 & 0.50 \\
O3 & 0.50 & 0.50 & \textbf{0.25} & 0.50 & 0.50 \\
O4 & 0.50 & 0.50 & 0.50 & \textbf{0.85} & 0.50 \\
O5 & 0.50 & 0.50 & 0.50 & 0.50 & \textbf{0.90} \\
\bottomrule
\end{tabular}
\end{table}

We design the examiner population using an orthogonal-plus-composite structure that balances causal interpretability with realistic profile representation. All examiners share four analytical skills grounded in U.S. patent law (\S\S102, 103, 101, 112); details in Appendix~\ref{app:prompts}. Their differences come from the five dimension values defined above and, for the two composite archetypes only, a legal-style label. The remaining persona fields are held at the same neutral setting for every archetype (Appendix~\ref{app:algorithm}).

\textit{Orthogonal archetypes.} Five archetypes (Table~\ref{tab:archetypes}) each set one dimension to an extreme value while holding the others at the 0.50 baseline. This design allows clean attribution of outcome differences to individual dimensions, free from the collinearity that arises when dimensions co-vary in real examiner populations.

Extreme values are calibrated to approximately the 85th--90th percentile of documented examiner variation (e.g., strictness 0.85 reflects the upper quartile of individual grant-rate variation reported in \citealt{lemley2012examiner}).

\textit{Composite archetypes.} Two archetypes represent documented real-world profiles in which dimensions naturally co-vary: C1 (veteran strict: 0.80/0.70/0.30/0.85/0.70, per \citealt{frakes2017time}) and C2 (lenient advocate: 0.30/0.35/0.70/0.40/0.60, per \citealt{farremensa2020patent}). These test whether multi-trait profiles produce emergent effects beyond what single-dimension analysis predicts (Section~\ref{sec:dimension-effects}). From each of the 7 archetypes, two instances are generated with Gaussian noise ($\sigma = 0.05$), yielding 14 examiners.

\section{Experiments}
\label{sec:experiments}

\subsection{Data}

\textit{Case pool.} PILOT-Bench \citep{jang2025pilot} provides 18,049 PTAB appeal decisions (2009--2024) but only the patent numbers, not the underlying specifications. We construct a full-text corpus by retrieving and parsing the USPTO publications for the applicant and prior-art patents, yielding a source pool of 7,698 cases with complete text for both. From this pool we sample our experimental cohort, each case carrying complete patent text (average 19.4 claims, 57K characters of specification, and full prior art). Filtering details are in Appendix~\ref{app:cohort}; the preprocessed dataset and pipeline code are released in our public repository (Appendix~\ref{app:fullresults}).

\textit{Stratified selection.} We draw a stratified sample of 100 cases enriched with boundary cases: 45 Affirmed, 35 Reversed, 15 Affirmed-in-Part, and 5 with new grounds of rejection (1 Affirmed, 2 Reversed, 2 Affirmed-in-Part). This yields 54 boundary cases in which PTAB reversed or partially reversed the examiner, over half the evaluation set, maximizing statistical power for disagreement analysis. We also prioritize multi-issue cases (58 cases) for cross-statute analysis. Full stratification details and selection bias checks are in Appendix~\ref{app:cohort}.

\subsection{Experiment and Metrics}

Our full 100-case experiment uses Qwen3-235B-A22B \citep{qwen2025qwen3} as the backbone, selected for its combination of frontier capability and open-weight reproducibility. We run 100 cases $\times$ 14 examiners = 1{,}400 decisions at temperature = 0.4, seed = 42, JSON structured output. Full implementation details for all experiments, including the cross-model and institutional-oversight runs, are in Appendix~\ref{app:archetype}; cross-model replication on five additional backbones is reported in Section~\ref{sec:robustness}.

\textit{Validation criteria.} We validate the judgment space through the three properties introduced in Section~\ref{sec:design}: (1) \textit{coverage}, the fraction of cases where the PTAB outcome direction is represented by at least one simulated examiner; (2) \textit{profile-attributability}, whether disagreements trace to specific profile dimensions, assessed in Section~\ref{sec:coverage}; and (3) \textit{legal coherence}, assessed for each examiner trace on a 5-item coherence checklist covering statute-appropriate framework, evidence-linked reasoning, stage consistency, dimension-consistent reasoning, and non-arbitrary final decision (full definitions in Appendix~\ref{app:cascade}).

\section{Results and Analysis}
\label{sec:results}

We evaluate the generated judgment space against the three validation properties from Section~\ref{sec:design}: coverage, profile-attributability, and legal coherence. Across 1{,}400 decisions (100 cases $\times$ 14 examiners; per-case matrix in Appendix~\ref{app:fullresults}), all three hold. We then analyze \textit{why} examiners diverge: which dimensions drive disagreement (\S\ref{sec:dimension-effects}), where in the pipeline it concentrates (\S\ref{sec:process}), and how institutional oversight reshapes it (\S\ref{sec:oversight}).

\begin{figure}[t]
\centering
\includegraphics[width=\columnwidth]{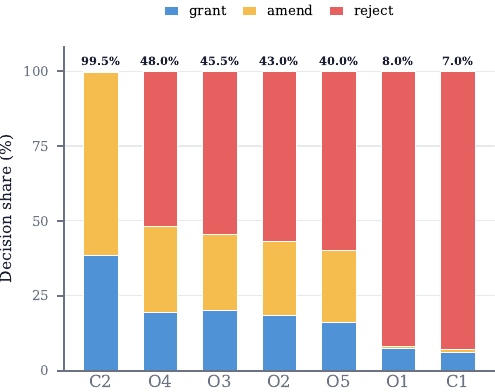}
\caption{Per-archetype decision distribution across 1{,}400 decisions (100 cases $\times$ 14 examiners). Stacked bars show grant / amend / reject share by archetype, ordered left-to-right from most lenient (C2) to most strict (C1). Numbers above each bar indicate the grant-plus-amend (favorable) rate. Archetype legend: \textbf{C2} = lenient advocate, \textbf{O4} = senior examiner, \textbf{O3} = easy combiner, \textbf{O2} = broad interpreter, \textbf{O5} = domain specialist, \textbf{O1} = strict gatekeeper, \textbf{C1} = veteran strict.}
\label{fig:archetype_bar}
\end{figure}

\subsection{Structured Disagreement}

As shown in Figure~\ref{fig:archetype_bar}, the grant-plus-amend rate across archetypes spans 92.5 percentage points: the lenient composite C2 reaches 99.5\% (38.5\% grant, 61.0\% amend), four orthogonal single-dimension archetypes occupy the moderate range (40--48\%), and the two strict archetypes anchor the low end (O1 strict gatekeeper 8.0\%, C1 strict composite 7.0\%). The spread between C2 and C1/O1 captures the ``patent lottery'' range documented in the empirical literature \citep{lemley2012examiner}.

This diversity is not an artifact of LLM stochasticity or persona labeling (Appendix~\ref{app:baseline}), and the lenient and strict archetypes keep their rank across additional LLM backbones (\S\ref{sec:robustness}).

\subsection{Dimension Effects}
\label{sec:dimension-effects}

\textit{Strictness dominates.} Examination strictness is by far the most influential dimension among the five orthogonal archetypes.

The baseline grant-plus-amend rate (the directly measured rate at all dimensions = 0.50, $n=100$ cases $\times$ 14 neutral examiners) is 41.6\%. Raising strictness alone to 0.85 pulls the grant+amend rate to 8.0\%, an effect of $-$33.6 pp, more than five times larger than any other dimension. The direction of this effect is designed (\S\ref{sec:design}). Its size, and the gap to the other four dimensions, is not.

The remaining four dimensions produce smaller effects within a $\pm$7 pp range: seniority $+$6.4 pp, combination threshold $+$3.9 pp, claim breadth $+$1.4 pp, and tech specialization $-$1.6 pp. Seniority is the clearest case. We gave it no direction in the prompt, yet senior examiners grant more, as \citet{frakes2017time} report for real examiners. Tech specialization shows a small negative direction consistent with \citet{righi2019examiner}, though the effect is weaker than reported there.

\textit{Super-additive interaction.} The composite archetypes show an asymmetry in how dimensions combine.

C2 (lenient advocate) lands at 99.5\% grant+amend, more than double the highest single-dimension rate (48.0\%). When multiple dimensions are simultaneously configured in the lenient direction, their effects compound super-additively. An additive model that sums orthogonal effects scaled by each dimension's deviation from baseline would estimate roughly 55\% for C2's profile (Appendix~\ref{app:archetype}); the observed 99.5\% substantially exceeds this prediction by $+$44.8 pp.

C1 (veteran strict) lands at 7.0\% grant+amend, essentially identical to O1 (strict gatekeeper, 8.0\%) despite C1 having additional extreme dimensions (breadth 0.70, seniority 0.85, tech specialization 0.70). The additive model predicts 22.5\% for C1, but the observed 7.0\% falls 15.5 pp below this prediction. Strictness 0.80 already pushes the rate near floor, and the other extreme dimensions cannot lift it back up. In effect, high strictness overrides the influence of all other dimensions.

This pattern, in which strict dimensions dominate while lenient dimensions compound, is direct process-level evidence that the patent lottery is structurally asymmetric. A strict examiner produces rejections regardless of other characteristics. A lenient outcome requires alignment across multiple dimensions at once.

The orthogonal design therefore supports a statement that is difficult to make with observational data. In the simulated population, strictness alone, isolated from all other examiner characteristics, is enough to push outcomes to the rejection floor.

\subsection{Judgment Space Coverage and Explainability}
\label{sec:coverage}

We assess the two structural validation properties from Section~\ref{sec:design}: does the judgment space cover real institutional outcomes (coverage), and do disagreements trace to specific examiner characteristics (profile-attributability)?

\textit{Coverage.} In 99 of the 100 cases, the PTAB outcome direction is represented by at least one member of the examiner population. For affirmed cases, on average 61.3\% of examiners agree with the PTAB direction (reject); for reversed cases, 43.4\% agree (grant or amend). In those 99 cases the PTAB position sits inside the generated judgment space, as a majority position in affirmed cases and a substantial minority in reversed cases. The single miss is examined next.

\textit{The coverage gap.} The sole miss (case 2021004068) is a TC 3600 business method patent affirmed on \S101 grounds; no examiner flagged the claims as patent-ineligible. This pattern is consistent with PTAB \S101 affirmance rates exceeding 95\% for business method patents \citep{borella2025ptab} and is discussed further under Limitations.

Coverage alone is weak: a sufficiently diverse population trivially covers all outcomes; the real question is whether this coverage is structured.

The first piece of evidence is a manipulation check. Examiner strictness and the favorable-decision rate are strongly negatively correlated (Pearson $r = -0.957$ across the seven archetype means and $r = -0.948$ across the 14 individual examiners, both $p < 0.001$). The sign is fixed by design (\S\ref{sec:design}). What the correlation adds is that the backbone followed its parameters almost perfectly, at the archetype level and for each noise-perturbed instance. Ordering the archetypes by strictness reverses their ordering by favorable rate, with one exception at the strict end, where C1 and O1 differ by 1 pp. The judgment space is not a noisy cloud. Instead, it is graded by the very parameter we manipulated, with the favorable rate falling predictably as an examiner moves from lenient to strict. The disagreement is a direct response to the design dimensions we control, not a side effect of LLM variation.

\textit{Profile-attributable disagreement.} When examiners disagree with PTAB, the disagreement traces to specific profile dimensions, not only in selected examples but as the modal pattern across all reversed cases. Across the three reversed \S103 cases with the richest variation, PTAB-aligned examiners (grant/amend) average strictness 0.39--0.46 vs.\ 0.67--0.84 for PTAB-opposing ones (reject), which are almost always O1 and C1. In case 2017005064, for instance, the lenient examiner (strictness=0.25) grants on the same ``not taught by prior art'' reasoning the PTAB used, while the strict examiner (strictness=0.89) rejects on the same ``obvious extension'' reasoning the original USPTO examiner used.

Crucially, the strict examiners are not ``wrong'': both positions are legitimate readings under \textit{KSR v.~Teleflex}, and the gap between them is simply where each draws the line for motivation to combine, a single profile dimension.

This pattern holds across all 37 reversed cases, including 2 with new grounds of rejection. In every case, the reject subgroup has higher average strictness than the grant subgroup (permutation test $p < 0.0001$).

\textit{Legal coherence.} Both the lenient and strict positions in the case above satisfy all five items of our coherence checklist. To verify this holds across the full experiment, we score all 1{,}400 traces via LLM-as-judge \citep{zheng2023judging} with two judges, Claude Sonnet 4.6 and GPT-5.6 Sol. Under Sonnet 4.6, the mean score is 4.79/5 and 97.9\% of traces score 4--5/5 (classified legally coherent), with 81.0\% scoring 5/5. GPT-5.6 Sol is stricter, with a mean of 3.93/5 and 70.1\% at 4--5/5, but it lowers every archetype alike. For both judges the strongest item is evidence-linked reasoning (100\% and 99.4\% pass). The weakest item under Sonnet 4.6 is stage consistency (83.1\%), where later-stage conclusions occasionally extend earlier findings without re-stating the justification. GPT-5.6 Sol additionally fails many traces on the statute-appropriate framework item (44.9\%), mostly for novelty findings that borrow obviousness reasoning. Full coherence-scoring methodology, per-item pass rates, the comparison of the two judges, permutation-test procedures, and worked traces are detailed in Appendix~\ref{app:cascade}.

\subsection{Process-Level Analysis}
\label{sec:process}

\begin{figure}[t]
\centering
\begin{subfigure}[t]{\columnwidth}
\includegraphics[width=\columnwidth]{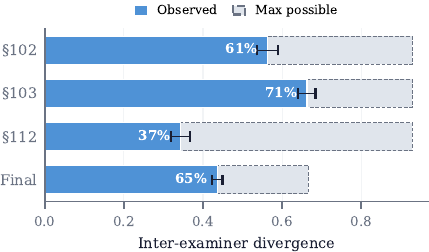}
\caption{Per-stage divergence vs.\ theoretical maximum.}
\label{fig:cascade-a}
\end{subfigure}\\[4pt]
\begin{subfigure}[t]{\columnwidth}
\includegraphics[width=\columnwidth]{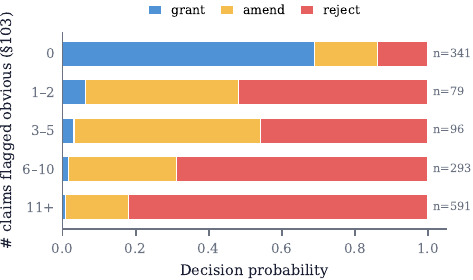}
\caption{Within-examiner: \S103 count $\to$ decision.}
\label{fig:cascade-b}
\end{subfigure}
\caption{Cascade structure of inter-examiner disagreement (100 cases, 1{,}400 decisions). (a) Per-stage divergence normalized by each stage's theoretical maximum; bar labels report this observed-to-ceiling ratio rather than the raw x-axis value. \S103 reaches 71\% of ceiling, well above the more factual §112 stage at 37\%. (b) Within-examiner relation between \S103 reject-flag count and the final decision: 0 flags yield 69\% grant, 11+ flags yield 82\% reject. \S103 is both where examiners diverge and where the divergence translates into outcomes.}
\label{fig:cascade}
\end{figure}

\textit{Disagreement concentrates at \S103.} Each pipeline stage has a different theoretical maximum divergence rate, set by its number of decision categories. Normalizing by this stage-specific ceiling (Figure~\ref{fig:cascade-a}), \S103 stands out as the analytical peak, well above the more factual \S112 stage \citep{knappich2025pedantic}. This is consistent with the legal observation that the ``motivation to combine'' standard under \S103 is the most discretionary point of the examination \citep{mandel2007patently}. The cohort is \S103-heavy (90/100 cases), so this concentration partly reflects the case mix; however, the per-stage normalization in Figure~\ref{fig:cascade-a} controls for stage frequency, so the peak reflects discretion density rather than case counts.

\textit{\S103 outcomes drive the final decision.} The number of claims an examiner flags as obvious at \S103 is a near-monotone predictor of the final decision (Figure~\ref{fig:cascade-b}). Disagreement that emerges at \S103 cascades into divergent outcomes rather than averaging out at the synthesis stage.

\begin{table}[h]
\centering
\caption{Cross-model robustness on a stratified 50-case subset ($n=50$). $r_G$: per-archetype Pearson correlation of grant rates with the main Qwen3-235B-A22B experiment. $\sigma^2$: within-model variance of archetype grant rates. $L_G$, $L_A$: lenient\_advocate grant and amend rates. Across all six models, the lenient composite stays highest and the strict pair lowest, and only the moderate archetypes reorder.}
\label{tab:robustness}
\small
\begin{tabular}{@{}lcccc@{}}
\toprule
\textbf{Model} & \textbf{$r_G$} & \textbf{$\sigma^2$} & \textbf{$L_G$} & \textbf{$L_A$} \\
\midrule
Qwen3-235B (main)  & ---    & 0.0098 & 0.385 & 0.610 \\
\midrule
Qwen3-8B           & 0.926  & 0.0008 & 0.090 & 0.530 \\
Qwen3-32B          & 0.901  & 0.0022 & 0.160 & 0.660 \\
Claude Sonnet 4.5  & 0.945  & 0.0125 & 0.400 & 0.540 \\
DeepSeek-V3        & 0.958  & 0.0319 & 0.570 & 0.410 \\
Mistral Large      & 0.879  & 0.0155 & 0.370 & 0.630 \\
\bottomrule
\end{tabular}
\end{table}

\subsection{Robustness across Model Families and Sizes}
\label{sec:robustness}

Our main findings rely on a single LLM backbone (Qwen3-235B-A22B). To test whether the structured judgment space arises from the design rather than from this specific model, we replicate the experiment on a stratified 50-case subset, matching the outcome distribution of the main 100-case cohort, using five additional models: Qwen3-8B, Qwen3-32B, Claude Sonnet 4.5, DeepSeek-V3, and Mistral Large. The five models span four frontier laboratories and four capability tiers from 8B parameters to frontier-closed.

\textit{Archetype-level grant rates correlate strongly with the main experiment.} As shown in Table~\ref{tab:robustness}, per-archetype grant rates across all five replications are highly correlated with the main 100-case Qwen3-235B-A22B results. The archetype ordering is preserved at both ends in all six models. lenient\_advocate has the highest favorable rate and strict\_gatekeeper and veteran\_strict the two lowest, while the four moderate archetypes sit between them in an order that varies by model (Figure~\ref{fig:crossmodel-heatmap}).

\textit{Calibration differs by family, the ordering at the extremes does not.} Within Qwen, smaller models compress the persona signal into \textit{amend} rather than grant ($\sigma^2$ scales with size). Across families, DeepSeek-V3 calibrates the lenient pole more strongly. Mistral Large moves nearly every decision outside lenient\_advocate to amend, so its grant rates are uniformly lower than the main experiment while its favorable rates are uniformly higher, and this compression explains its lower per-archetype correlation ($r_G = 0.879$ vs.\ $\geq 0.90$ elsewhere). The extremes nonetheless keep their rank in every replication, so the structured judgment space depends on the design, not the backbone.

\begin{figure}[t]
\centering
\includegraphics[width=\columnwidth]{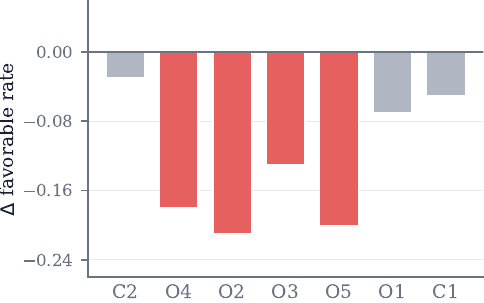}
\caption{Effect of institutional oversight on per-archetype favorable rate ($\Delta$ = treatment $-$ control). Archetypes ordered by baseline favorable rate (left = most lenient, right = most strict). The four moderate archetypes (O4, O2, O3, O5) show the largest reductions. The extremes (C2, O1, C1) move less in absolute terms, with the strict pair near the floor and C2 near the ceiling. Per-archetype values and $p$-values are in Appendix~\ref{app:oversight-full}.}
\label{fig:oversight}
\end{figure}

\subsection{Institutional Oversight Effects}
\label{sec:oversight}

Real patent examiners do not decide in isolation. Beyond their personal calibration, institutional constraints, including SPE signature review, OPQA quality audits, and asymmetric career consequences for erroneous allowance vs.\ erroneous rejection,
shape every decision. A natural question follows: how much does institutional oversight modulate the judgment space, and does it affect every examiner profile equally?

We test this by adding an oversight-awareness component to the persona block (full prompt in Appendix~\ref{app:prompts}) and re-running the same 50-case stratified subset under two conditions: \textit{control} (oversight = 0, equivalent to the main experiment) and \textit{treatment} (oversight = 1, full institutional framing). Each archetype is run identically in both conditions; the only difference is whether examiners read the oversight component before deciding.

\textit{Selective, non-uniform effect.} Adding institutional oversight does not push every archetype toward strictness by the same amount. Figure~\ref{fig:oversight} shows a U-shape. The four moderate archetypes lose 13 to 21 pp of favorable rate, three of them at $p < 0.01$. The three extremes move by 3 to 7 pp (C2 $p = 0.25$, C1 $p = 0.27$, O1 $p = 0.016$).

\textit{Floor and ceiling.} The small absolute shifts at the extremes have different sources. The strict archetypes start at 9--13\% favorable and have little room left to fall. C2 starts at 100\% and could fall the full way, yet still ends favorable in 97\% of cases. The moderate middle loses favorable rate mostly through an amend $\to$ reject migration (per-archetype shifts in Appendix~\ref{app:oversight-full}).

This complicates the policy intuition that ``more oversight'' uniformly reduces grant rates. In the simulated population, oversight reshapes the judgment space unevenly. It moves the moderate middle most in absolute terms, which is where the borderline cases sit, while the strict extremes are already near the floor and the lenient extreme hardly moves.

\section{Conclusion}
\label{sec:conclusion}

We address the single-sample constraint of patent examination research by generating, for each case, the full \textit{judgment space} of professional positions a population of differently calibrated examiners would produce. Across 100 cases $\times$ 14 examiners (1{,}400 decisions), the space covers 99/100 real institutional outcomes; disagreements trace to specific examiner dimensions with an asymmetric structure (strict dimensions override others, lenient outcomes require multi-dimensional alignment); patterns replicate across five additional LLM backbones; and institutional oversight reshapes the space unevenly. Which examiner a case lands with is a lottery, but the outcomes are not purely random: beneath the apparent chance, the judgment space reveals a structured, asymmetric pattern. The framework's value is not single-decision prediction but making the range of defensible professional positions visible and traceable: a foundation for pre-filing risk assessment and policy simulation that single-sample data cannot support.

\section{Limitations}
\label{sec:limitations}

\textit{Scope.} Our 100-case cohort with 14 examiners supports structural pattern analysis but limits fine-grained subgroup work (1--22 cases per TC; 8 of 10 TCs represented, with TC 1600 represented by only 1 case). The cohort is also boundary-enriched (54\% Reversed + AiP) for statistical power on disagreement \textit{structure} analysis; disagreement \textit{prevalence} estimates should not be extrapolated to the broader PTAB population, where Affirmed cases dominate ($\sim$60\%).

\textit{\S101 eligibility.} The pipeline has no dedicated \S101 stage. Eligibility enters only as a persona-level instruction to flag claims that may recite an abstract idea, and the examiners almost never act on it. Of the 29 cohort cases whose PTAB decision involved \S101, the examiners raised \S101 in one (recall 1/29), and in TC 3600, where 17 of our 21 cases involve \S101, recall is 0/17. The coverage miss in \S\ref{sec:coverage} is one visible symptom. Since PTAB \S101 affirmance rates exceed 95\% for business method patents \citep{borella2025ptab}, deployment in \S101-dominant technology areas is premature. Adding a dedicated \textit{Alice}/\textit{Mayo} stage is the most important next step.

\textit{Coherence judge.} The checklist measures the internal coherence of a trace, whether it applies the right statute, ties its conclusions to evidence, and stays consistent across stages. It does not measure whether the legal conclusions are correct. The scores come from two LLM judges that set different bars, GPT-5.6 Sol failing 392 traces that Sonnet 4.6 passes (Appendix~\ref{app:llm-judge}), and neither judge has been validated against human experts. Recruiting patent professionals across the eight technology centers in the cohort was not realistic at this stage, so we fix the validation protocol in advance (Appendix~\ref{app:human-protocol}). Until it is run, the coherence figures in \S\ref{sec:coverage} should be read as the judgment of two LLMs with different thresholds, not as a validated measure.

\textit{Temporal context.} The framework operates in a single fixed temporal context. Examiner judgments also depend on the prevailing doctrine, which \textit{Alice Corp.} (2014) reshaped for \S101 and \textit{KSR v.~Teleflex} (2007) for \S103, and on the state of ordinary skill in the art, which evolves across technology generations. What was non-obvious to a software engineer in the 1990s may be obvious to one in 2020. Comparing judgment spaces across policy eras and PHOSITA states is a natural extension of the framework.

\textit{Methodological limitations.} Our finding that \S112 shows high convergence may partly reflect simulation limitations rather than genuine low discretion (PEDANTIC; \citealt{knappich2025pedantic}). Orthogonal archetypes do not represent realistic profiles (no real examiner differs on exactly one dimension); the two composite archetypes partially address this but cannot span the full space.

\section*{Ethical Considerations}

\textit{Data and human subjects.} PTAB decisions and the underlying patent documents are public USPTO records, and the PILOT-Bench corpus we build on is publicly available. The study involves no human subjects and no personally identifiable information, so no IRB review or annotator compensation was required.

\textit{Intended use, dual use, and reproducibility.} Our framework generates a distribution of professional positions to make the range of defensible examination outcomes visible; it is a research instrument for studying examiner disagreement, not a system for rendering or replacing real patent decisions. As noted in Limitations, the simulated examiners do not match human-examiner accuracy, and their outputs should not be used to grant, reject, or pre-judge any application; the main foreseeable misuse is over-reliance on simulated outcomes in place of qualified legal judgment. Dual-use risk is otherwise low. Models, parameters, seeds, and prompts are reported in the paper and appendices, and code and data are released in the public repository (Appendix~\ref{app:fullresults}). AI assistants were used for coding support and for language and typo editing, not for research ideation or analysis.

\section*{Acknowledgments}
This research was supported by the MSIT (Ministry of Science and ICT), Korea, under a grant through the National Research Foundation of Korea (NRF) (No.~RS-2024-00354675, 33.3\%). This research was also supported by the MSIT, Korea, under the ICAN (ICT Challenge and Advanced Network of HRD) support program supervised by the IITP (Institute for Information \& Communications Technology Planning \& Evaluation) (IITP-2023-RS-2023-00259806, 33.3\%). This work was supported by the National Research Foundation of Korea (NRF) grant funded by the Korea government (MSIT) (No.~RS-2026-25518141, 33.3\%).
 
\bibliography{references}

\clearpage

\appendix


\section{Archetype Parameter Details}
\label{app:archetype}

\textit{Noise configuration.} Each dimension receives independent Gaussian noise ($\sigma = 0.05$), clipped to $[0, 1]$. One seeded generator per instance draws the noise in a fixed order, and a neutral policy-stance field (base $0.50$, $\sigma = 0.02$) is drawn second, between strictness and breadth. Reproducing Table~\ref{tab:realized} from the seeds requires this draw (Algorithm~\ref{alg:jspace}). Seeds start at 42 and increase by 1 per instance in sorted archetype-file order. The realized parameter values for all 14 examiner instances are listed in Table~\ref{tab:realized}.

\begin{table*}[t]
\centering
\caption{Realized examiner parameters (14 instances). Bold values in the main text tables indicate the isolated extreme dimension for each orthogonal archetype. Noise produces small perturbations (typically $\pm$0.05) around base values, ensuring that the two instances of each archetype are similar but not identical.}
\label{tab:realized}
\small
\begin{tabular}{@{}clcccccc@{}}
\toprule
\textbf{\#} & \textbf{Archetype} & \textbf{ID} & \textbf{Strict.} & \textbf{Breadth} & \textbf{Comb.} & \textbf{Senior.} & \textbf{Spec.} \\
\midrule
1  & C1: Veteran strict     & 0042 & 0.793 & 0.694 & 0.335 & 0.844 & 0.625 \\
2  & C1: Veteran strict     & 0043 & 0.875 & 0.734 & 0.344 & 0.808 & 0.622 \\
3  & C2: Lenient advocate   & 0044 & 0.248 & 0.370 & 0.676 & 0.402 & 0.612 \\
4  & C2: Lenient advocate   & 0045 & 0.292 & 0.390 & 0.722 & 0.407 & 0.603 \\
5  & O1: Strict gatekeeper  & 0046 & 0.889 & 0.416 & 0.448 & 0.508 & 0.560 \\
6  & O1: Strict gatekeeper  & 0047 & 0.818 & 0.456 & 0.513 & 0.450 & 0.496 \\
7  & O2: Broad interpreter  & 0048 & 0.475 & 0.790 & 0.459 & 0.530 & 0.488 \\
8  & O2: Broad interpreter  & 0049 & 0.547 & 0.864 & 0.553 & 0.484 & 0.499 \\
9  & O3: Easy combiner      & 0050 & 0.461 & 0.476 & 0.222 & 0.369 & 0.522 \\
10 & O3: Easy combiner      & 0051 & 0.502 & 0.549 & 0.329 & 0.507 & 0.563 \\
11 & O4: Senior examiner    & 0052 & 0.517 & 0.491 & 0.443 & 0.898 & 0.482 \\
12 & O4: Senior examiner    & 0053 & 0.422 & 0.424 & 0.521 & 0.906 & 0.477 \\
13 & O5: Domain specialist  & 0054 & 0.523 & 0.446 & 0.480 & 0.547 & 0.893 \\
14 & O5: Domain specialist  & 0055 & 0.603 & 0.548 & 0.566 & 0.545 & 0.858 \\
\bottomrule
\end{tabular}
\end{table*}

\textit{Excluded dimensions.} We considered but excluded three additional dimensions: \textit{policy stance} (better modeled as an environment-level variable; see Section~\ref{sec:limitations}), \textit{written description rigor} (\S112 shows high convergence; Section~\ref{sec:process}), and \textit{industry background} (would compromise the orthogonal design by introducing collinearity with technology specialization).

\begin{figure}[t]
\centering
\includegraphics[width=\columnwidth]{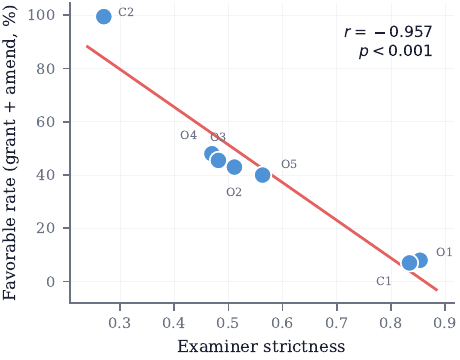}
\caption{Examiner strictness vs.\ favorable (grant + amend) rate across the seven archetypes. Each point is one archetype's mean over 200 decisions (100 cases $\times$ 2 instances). Ordering by strictness reverses the ordering by favorable rate except for the two strictest archetypes, C1 and O1, which differ by 1 pp (Pearson $r = -0.957$, $p < 0.001$), so the judgment space is graded by the manipulated strictness parameter rather than dispersed as noise.}
\label{fig:strictness-scatter}
\end{figure}

The strong monotonic relationship between strictness and favorable rate is shown in Figure~\ref{fig:strictness-scatter}.

\begin{figure}[t]
\centering
\includegraphics[width=\columnwidth]{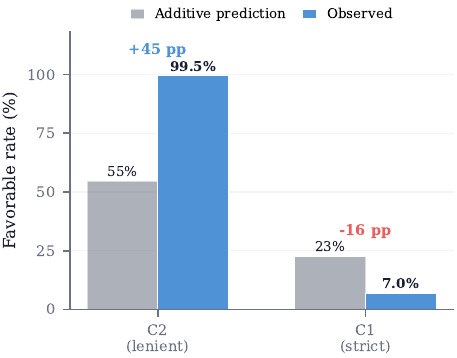}
\caption{Additive prediction (grey) vs.\ observed favorable rate (blue) for the two composite archetypes. The additive model sums each orthogonal dimension's marginal effect scaled by its deviation from the 0.50 baseline. C2 (lenient advocate) exceeds its prediction by $+$45 pp --- lenient dimensions compound super-additively --- whereas C1 (veteran strict) falls 16 pp below, because a single extreme strictness value already saturates the rate near the rejection floor and leaves no room for the other extreme dimensions to act. Values are rounded for display; the additive predictions and gaps are 55\%/$+$44.8 pp for C2 and 22.5\%/$-$15.5 pp for C1 (Appendix~\ref{app:archetype}).}
\label{fig:additive-observed}
\end{figure}

\textit{Additive prediction for composite archetypes.} The additive baseline against which super-additive effects in Section~\ref{sec:dimension-effects} are measured is computed by treating each dimension's effect as independent and summing scaled contributions. For a composite profile with dimension values $\{x_d\}_{d \in \text{dims}}$, the predicted grant+amend rate is:
\[
\hat{p}_{\text{additive}} = p_{\text{baseline}} + \sum_{d} \beta_d \cdot (x_d - 0.50)
\]
where $\beta_d$ is the per-dimension slope estimated from the orthogonal archetypes (linear extrapolation from the (0.50, $p_{\text{baseline}}$) to (extreme value, observed rate) point for each dimension), with $p_{\text{baseline}} = 41.6\%$ measured directly from a 14-examiner all-dims-0.50 control on the same 100-case cohort. Applying this to C2 (lenient advocate, 0.30/0.35/0.70/0.40/0.60) yields $\hat{p}_{\text{additive}} \approx 55\%$ vs.\ the observed 99.5\%, a $+$44.8 pp super-additive deviation. C1 (veteran strict, 0.80/0.70/0.30/0.85/0.70) yields $\hat{p}_{\text{additive}} \approx 22.5\%$, but the observed 7.0\% falls 15.5 pp below the prediction. Strictness 0.80 already saturates the rate near the rejection floor, and additional extreme dimensions in seniority, breadth, and tech specialization cannot lift it back. This asymmetric overshoot is direct evidence that strict-dimension dominance is non-linear. It caps the floor more aggressively than an additive model anticipates (Figure~\ref{fig:additive-observed}).

\begin{figure}[t]
\centering
\includegraphics[width=\columnwidth]{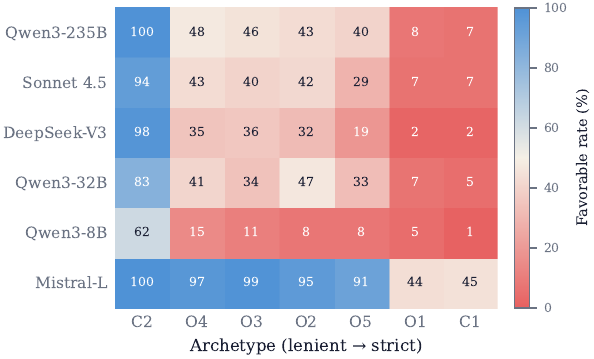}
\caption{Per-archetype favorable rate across the six LLM backbones (rows) and seven archetypes (columns, ordered lenient $\to$ strict). In every backbone, C2 remains highest and O1 and C1 lowest, while the four moderate archetypes between them shift in order from model to model. Absolute calibration differs by family. Qwen3-8B is uniformly more conservative and Mistral Large uniformly more lenient in favorable rate. The structural finding is thus separable from model-specific calibration.}
\label{fig:crossmodel-heatmap}
\end{figure}

\textit{Cross-model setup.} The robustness replication in Section~\ref{sec:robustness} uses a stratified 50-case subset of the main 100-case cohort, drawn with seed = 42 and quotas that preserve the source outcome distribution (23 Affirmed, 18 Reversed, 7 Affirmed-in-Part, 2 Other with new grounds of rejection); the same subset is reused for the oversight experiment below. Each replication runs the identical 14-examiner population (7 archetypes $\times$ 2 instances, base seed = 42), examination pipeline (the 8 stages defined in Appendix~\ref{app:pipeline}), and inference settings (temperature = 0.4, max\_tokens = 16{,}000, 300\,s per-call timeout, JSON structured output) as the main experiment; only the LLM backbone changes. The five backbones are: Qwen3-8B (\texttt{Qwen/Qwen3-8B}, served locally via a HuggingFace endpoint), Qwen3-32B (\texttt{qwen/qwen3-32b} via OpenRouter), Claude Sonnet 4.5 (\texttt{claude-sonnet-4-5} via the Anthropic API directly), DeepSeek-V3 (\texttt{deepseek/deepseek-chat-v3-0324} via OpenRouter), and Mistral Large (\texttt{mistralai/mistral-large-2512} via OpenRouter). All API runs use four concurrent worker threads. Figure~\ref{fig:crossmodel-heatmap} shows the per-archetype favorable rate for all six backbones.

\textit{Institutional-oversight setup.} The oversight experiment in Section~\ref{sec:oversight} runs on the same 50-case subset and the same 14-examiner population, with two conditions on the \texttt{oversight\_awareness} parameter --- control (\texttt{oversight\_awareness} = 0.0, no oversight component injected into the persona prompt) and treatment (\texttt{oversight\_awareness} = 1.0, oversight component appended). The treatment component frames three institutional mechanisms in the persona block: \textit{SPE signature review} (allowances require Senior Patent Examiner sign-off; weakly grounded allowances may be returned), \textit{OPQA quality audit} (random audits flag missed rejection grounds as permanent quality defects), and \textit{asymmetric accountability} (erroneous rejections can be corrected on appeal, while erroneous allowances are recorded against the individual examiner). The model and all inference settings are held fixed at the main-experiment configuration (Qwen3-235B-A22B, \texttt{qwen/qwen3-235b-a22b-2507}, temperature = 0.4, base seed = 42); only the oversight component differs between conditions. Per-archetype favorable-rate shifts are tested with paired permutation (1{,}000{,}000 sign-flips, seed = 42) over the 50 shared cases; results are reported in Table~\ref{tab:oversight-full}.

\begin{figure*}[t]
\centering
\includegraphics[width=\textwidth]{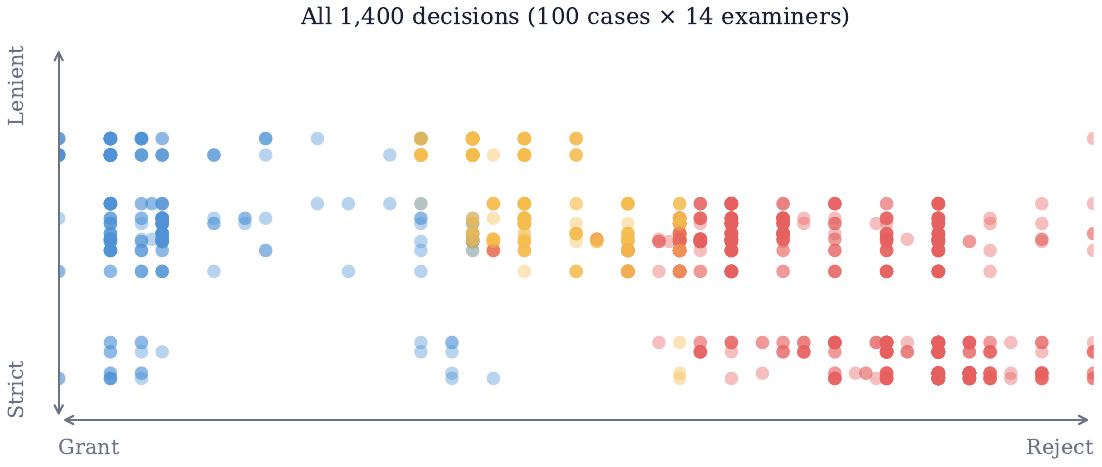}
\caption{All 1{,}400 decisions (100 cases $\times$ 14 examiners), placed by examiner strictness (vertical axis: lenient at top, strict at bottom) and decision direction (horizontal axis: grant at left, reject at right). Color encodes the decision (blue = grant, yellow = amend, red = reject). The decisions form a clear strictness gradient rather than a uniform cloud: lenient examiners concentrate on the grant side, strict examiners on the reject side, and amend decisions occupy the moderate middle. The horizontal banding reflects the seven archetypes' base strictness values, each with two noise-perturbed instances.}
\label{fig:judgment-overlay}
\end{figure*}

\section{Full Results}
\label{app:fullresults}

The per-case decision matrix (100 cases $\times$ 14 examiners, 1{,}400 decisions), the preprocessed 100-case full-text inputs and 50-case subset, per-case results for all experiments (main, baselines, oversight, cross-model, coherence scoring), and the analysis code are available at \url{https://github.com/YeHoonJang/patent-lottery}. The full-text patent corpus for the source pool (applicant and prior-art specifications) is available at \url{https://huggingface.co/datasets/Yehoon/ptab_patents}. The code is released under the MIT License and the corpus, which consists of public USPTO records, under CC BY 4.0.

Figure~\ref{fig:judgment-overlay} plots the entire decision set, showing that the judgment space is organized by examiner strictness rather than dispersed uniformly.

\section{Cascade Analysis Methodology}
\label{app:cascade}

This appendix details the cascade analysis referenced in Sections~\ref{sec:coverage} and~\ref{sec:process}. We describe the legal coherence checklist and its per-item scoring, the LLM-as-judge methodology used to score all 1{,}400 traces, and a worked case study; trace extraction, stage-boundary definitions, and the divergence scoring rubric follow the same pipeline stages defined in Appendix~\ref{app:pipeline}.

\subsection{Legal Coherence Checklist}

Each examiner trace is evaluated on five binary items (Table~\ref{tab:checklist}):

\begin{table*}[t]
\centering
\caption{Legal coherence checklist. Each item: 1 (pass) or 0 (fail). Traces scoring 4--5 are classified as legally coherent; 2--3 as partially coherent; 0--1 as weak.}
\label{tab:checklist}
\small
\begin{tabular}{@{}clp{4.8cm}p{4.8cm}@{}}
\toprule
\textbf{\#} & \textbf{Item} & \textbf{Pass criterion} & \textbf{Fail example} \\
\midrule
1 & Statute-appropriate framework & Invokes the recognized legal test for the relevant statute (e.g., Graham/KSR for \S103, element-by-element anticipation for \S102, Alice/Mayo for \S101, claim-specification support for \S112) & \S103 analysis that uses \S101-style abstract-idea reasoning instead of obviousness factors \\
2 & Evidence-linked reasoning & Conclusions reference specific claim language, prior art disclosures, or specification passages & ``This claim is obvious'' with no prior art citation or element comparison \\
3 & Stage consistency & Later-stage conclusions do not contradict earlier-stage findings without stated justification & Finding weak prior art overlap at \S102, then asserting strong \S103 rejection on the same elements without additional rationale \\
4 & Dimension-consistent reasoning & The examiner's profile dimensions are reflected in the reasoning (e.g., strict examiner resolves ambiguity toward rejection; easy combiner accepts weaker motivation to combine) & A strict examiner (strictness=0.85) granting without any stated weighing of ambiguous evidence \\
5 & Non-arbitrary final decision & Final decision follows from the accumulated stage-level analyses with stated rationale & Final decision that reverses the direction of all prior-stage conclusions without explanation \\
\bottomrule
\end{tabular}
\end{table*}

\textit{Worked example: Case 2017005064, \S103 disagreement.} The lenient examiner (strictness=0.25) scores 5/5: invokes Graham four-factor test (item 1), cites specific formula-based claim elements not found in prior art (item 2), maintains consistent ``novel feature'' finding from \S102 through \S103 (item 3), resolves ambiguity toward grant consistent with low strictness (item 4), and synthesizes a non-reject decision from accumulated non-obvious findings (item 5). The strict examiner (strictness=0.89) also scores 5/5: same Graham framework (item 1), cites overlapping LTE techniques in prior art (item 2), consistent ``prior art covers'' finding across stages (item 3), resolves ambiguity toward rejection consistent with high strictness (item 4), and synthesizes rejection from accumulated obvious findings (item 5). Both traces are legally coherent; the disagreement traces to where each examiner draws the combination threshold line.

\subsection{LLM-as-Judge Scoring Methodology}
\label{app:llm-judge}

We score all 1{,}400 traces from the main experiment on the 5-item checklist with two judges, Claude Sonnet 4.6 (Anthropic) and GPT-5.6 Sol (OpenAI) \citep{zheng2023judging}. Each judge receives the same prompt, operates on the concatenated stage outputs (examiner profile, \S102 novelty assessment, \S103 obviousness analysis, office action, and final decision), and returns a binary 0/1 score plus a one-sentence justification per item.

\textit{Per-item pass rates} across all 1{,}400 traces are reported in Table~\ref{tab:coherence-scoring}:

\begin{table}[h]
\centering
\caption{Per-item pass rates from LLM-as-judge scoring of the same 1{,}400 traces by two judges.}
\label{tab:coherence-scoring}
\small
\resizebox{\columnwidth}{!}{%
\begin{tabular}{@{}lcc@{}}
\toprule
\textbf{Item} & \textbf{Sonnet 4.6} & \textbf{GPT-5.6 Sol} \\
\midrule
1. Statute-appropriate framework & 99.6\% & 44.9\% \\
2. Evidence-linked reasoning & 100.0\% & 99.4\% \\
3. Stage consistency & 83.1\% & 61.1\% \\
4. Dimension-consistent reasoning & 98.1\% & 91.3\% \\
5. Non-arbitrary final decision & 97.7\% & 96.6\% \\
\bottomrule
\end{tabular}}
\end{table}

\textit{Score distribution.} Mean total score is 4.79/5 under Sonnet 4.6 and 3.93/5 under GPT-5.6 Sol. Under Sonnet 4.6, 81.0\% of traces score 5/5, 16.9\% score 4/5, 1.9\% score 3/5, and 0.2\% score 2/5; no trace scores below 2. Under GPT-5.6 Sol, 27.2\% score 5/5, 42.9\% score 4/5, 25.9\% score 3/5, 3.9\% score 2/5, and 0.1\% score 1/5. The two judges agree on the 4-or-5 threshold in 71.7\% of traces (978 both pass, 26 both fail). Almost all disagreement runs one way. Sonnet 4.6 passes and GPT-5.6 Sol fails 392 traces, and the reverse happens in 4. GPT-5.6 Sol gives an equal or lower total in 98.0\% of traces, and the Pearson correlation between the two totals is 0.31.

\textit{Interpretation.} Evidence-linked reasoning is near-perfect under both judges (100\% and 99.4\%), which reflects that the structured pipeline grounds every analysis in specific claim language and prior-art content. The judges part ways on the statute-appropriate framework item (99.6\% vs.\ 44.9\%). In its justifications, GPT-5.6 Sol most often faults the \S102 stage for treating a combination of references, or an ``obvious variant,'' as anticipation, typically while crediting the Graham and KSR analysis at \S103. Stage consistency is the weakest item under Sonnet 4.6 (83.1\%) and the second weakest under GPT-5.6 Sol (61.1\%). In these traces, \S103 conclusions extend findings not explicitly noted at \S102 without re-stating the prior-art linkage. This is a soft inconsistency rather than a contradiction, since the additional reasoning is typically valid but not anchored back to the earlier stage. We treat it as a methodological observation rather than a coverage gap. The two judges therefore see the same weakness, novelty findings that borrow obviousness reasoning, and differ in how severely they penalize it. Per-archetype means range from 4.68 to 4.83 under Sonnet 4.6 and from 3.81 to 4.04 under GPT-5.6 Sol, so neither judge singles out any archetype, and the observation in \S\ref{sec:coverage} that lenient and strict positions are both coherent does not depend on the judge. Which judge is closer to a patent professional is the question the protocol in Appendix~\ref{app:human-protocol} is designed to answer.

\subsection{Protocol for Human Validation of the Coherence Judge}
\label{app:human-protocol}

The scores in Table~\ref{tab:coherence-scoring} come from two LLM judges that disagree on 28\% of traces. This section fixes the protocol for validating them against patent professionals (see Limitations for why it has not been run).

\textit{Evaluators.} Three evaluators with U.S. patent prosecution experience, either registered patent attorneys or agents or former USPTO examiners. Their combined expertise should span the mechanical, electrical, and software and business-method areas that make up most of the cohort, so that every sampled trace is read by at least one evaluator from a related field.

\textit{Sample.} A fixed set of 60 traces, scored by all three evaluators. Thirty are drawn from the 396 traces on which the two judges disagree on the 4-or-5 threshold, including all 4 that GPT-5.6 Sol passes and Sonnet 4.6 fails. Fifteen are drawn from the 26 traces that both judges fail, and 15 from the 978 that both judges pass. Each draw is stratified so that every archetype appears in each group.

\textit{Materials.} Evaluators receive what the judge received, namely the examiner profile and the \S102, \S103, office action, and final decision outputs, rendered as a readable document, together with the checklist in Table~\ref{tab:checklist}. They do not see either judge's scores or justifications, the other evaluators' scores, or the PTAB outcome.

\textit{Procedure.} A calibration session on five traces outside the sample, scored jointly and discussed until the item definitions are shared. Independent scoring then follows in randomized trace order, with one 0/1 mark and a one-line justification per item.

\textit{Analysis.} The analysis reports (i) agreement among the three evaluators per item, measured by Krippendorff's $\alpha$, (ii) agreement between the human majority vote and each judge, per item and on the 4-or-5 coherence threshold, and (iii) the direction of disagreement, that is, whether a judge passes items the humans fail more often than the reverse, and which judge the human majority sides with on the traces where the two judges disagree. Because the sample over-represents disputed traces, agreement is reported per stratum. The analysis plan is fixed before scoring begins.

\section{Representative Prompt Examples}
\label{app:prompts}

We provide representative excerpts from three key prompt components to show how examiner dimensions enter the prompts as numerical parameters rather than natural-language labels. The operational prompts are authored in a non-English language; the excerpts below are English renderings shown for readability. The verbatim original-language prompts for all skills and pipeline stages are available in the released repository (Appendix~\ref{app:fullresults}).

\textit{Section 103 Skill Prompt (excerpt).} Each examiner receives the same analytical framework, the Graham v.~John Deere four-factor test, but applies it through their profile-specific thresholds. The combination threshold enters as a numerical cutoff, not a label, which is what keeps the dimension stable across runs (in contrast to the persona prompts in Appendix~\ref{app:baseline}):

\begin{promptbox}
You are a USPTO patent examiner. Your role: obviousness assessment under 35 USC 103. Your combination_threshold = {examiner.combination_threshold}.

Apply the Graham v. John Deere four-factor analysis: Factor 1: Scope and content of the prior art. Factor 2: Differences between the prior art and the claims. Factor 3: Level of ordinary skill in the art (PHOSITA). Factor 4: Secondary considerations.

The lower your combination_threshold, the more readily you accept a motivation to combine prior art. If combination_threshold < 0.4: accept combination based on common-sense or design-choice reasoning. If combination_threshold > 0.6: require explicit teaching or suggestion in the prior art.
\end{promptbox}

\textit{Cognitive Framing (strictness-dependent).} Injected into the persona block to model decision framing \citep{tversky1981framing}. The numerical strictness value selects the framing clause deterministically, so the same value always yields the same framing:

\begin{promptbox}
[Strictness >= 0.7 -- Public-protection framing] “When the evidence is ambiguous, rejection serves the public interest. Rather than granting a marginal patent, it is better to protect only clearly valid inventions.”

[Strictness <= 0.4 -- Innovation-support framing] “When the evidence is ambiguous, the benefit of the doubt favors the applicant. Over-rejecting innovation is more harmful than over-granting it.”
\end{promptbox}

\textit{Weighing Step in Final Decision.} Added to the final synthesis to activate interpretive discretion. Strictness modulates only the ambiguous middle: strong grounds are maintained regardless of profile, so the parameter shifts outcomes without overriding the evidence:

\begin{promptbox}
Step 1: List all rejection grounds from previous stages. Step 2: Evaluate each ground's strength (strong / moderate / weak). Step 3: Given your strictness={examiner.strictness}: - Strong grounds: maintain rejection regardless of profile. - Moderate/weak grounds: decide based on your examination philosophy.
\end{promptbox}

\section{Examination Pipeline --- Legal and Procedural Grounding}
\label{app:pipeline}

Our 8-stage examination pipeline mirrors the standard USPTO patent examination process as codified in the Manual of Patent Examining Procedure (MPEP). This appendix documents the legal basis for each pipeline stage, each examiner dimension, and each analytical skill.

\subsection{Algorithmic Summary}
\label{app:algorithm}

Algorithm~\ref{alg:jspace} restates the procedure of \S\ref{sec:design} in the form in which it is implemented. \textsc{GeneratePopulation} runs once, and the 14 profiles it returns are reused for every case. \textsc{JudgmentSpace} builds $J(c)$ for one case by calling the shared pipeline $F$ (\textsc{ExecutePipeline}) once per profile. Every stage receives the same persona block $\Pi(\theta)$, which prints every profile field and, when strictness is $\geq 0.7$ or $\leq 0.4$, a public-protection or innovation-support framing. The two directional rules of \S\ref{sec:design} are attached to stages 4 and 8 only. The seeds shown are the ones used in all reported experiments, and the values they produce are listed in Table~\ref{tab:realized}.

\begin{algorithm*}[t]
\caption{Judgment space generation as implemented. $A_k$ is the base profile of archetype $k$ (Table~\ref{tab:archetypes}; C1 and C2 in \S\ref{sec:population}). Archetype files are read in sorted name order (C1, C2, O1, \ldots, O5), so seeds 42--43 belong to C1, 44--45 to C2, 46--47 to O1, and so on through 54--55 for O5.}
\label{alg:jspace}
\begin{algorithmic}[1]
\Procedure{GeneratePopulation}{$\{A_1,\ldots,A_7\}$, $s_0 = 42$}
  \State $\Theta \gets \emptyset$;\; $s \gets s_0$
  \For{$k = 1, \ldots, 7$ in sorted archetype-file order}
    \For{$j = 1, 2$}
      \State $\mathrm{rng} \gets \textsc{Random}(s)$
      \For{$\mathrm{dim} \in (\text{strictness}, \text{policy\_stance}, \text{breadth}, \text{combination}, \text{seniority}, \text{specialization})$}
        \State $\sigma \gets 0.02$ if $\mathrm{dim} = \text{policy\_stance}$, else $0.05$
        \State $\theta[\mathrm{dim}] \gets \mathrm{clip}\big(A_k[\mathrm{dim}] + \mathrm{rng}.\mathcal{N}(0,\sigma^2),\ 0,\ 1\big)$ \Comment{rounded to 3 decimals}
      \EndFor
      \State $\theta[\text{legal\_style}], \theta[\text{tech\_domains}] \gets A_k[\cdot]$ \Comment{no noise; textualist (C1), purposivist (C2), pragmatist (others); \texttt{general}}
      \State $\Theta \gets \Theta \cup \{\theta\}$;\; $s \gets s + 1$
    \EndFor
  \EndFor
  \State \Return $\Theta$ \Comment{$|\Theta| = 14$. policy\_stance is $0.50$ in every $A_k$ and is used by no stage rule}
\EndProcedure
\Statex
\Procedure{JudgmentSpace}{$c$, $\Theta$}
  \State $J \gets \emptyset$
  \For{$\theta_i \in \Theta$}
    \State $(d_i, T_i) \gets \textsc{ExecutePipeline}(c, \theta_i)$
    \State $J \gets J \cup \{(d_i, \theta_i, T_i)\}$
  \EndFor
  \State \Return $J$ \Comment{no aggregation}
\EndProcedure
\Statex
\Procedure{ExecutePipeline}{$c$, $\theta$} \Comment{each line below is one LLM call with system prompt $\Pi(\theta)$}
  \State $\mathrm{interp} \gets \textsc{ClaimInterpretation}(c)$ \Comment{breadth restated; no rule}
  \State $\mathrm{map} \gets \textsc{PriorArtMapping}(c, \mathrm{interp})$
  \State $\mathrm{nov} \gets \textsc{Novelty}_{\S102}(c, \mathrm{map})$ \Comment{strictness restated; no rule}
  \State $\mathrm{obv} \gets \textsc{Obviousness}_{\S103}(c, \mathrm{map}, \mathrm{nov})$ \Comment{rule: combination $<0.4$, common sense suffices; $>0.6$, explicit teaching required}
  \State $\mathrm{wd} \gets \textsc{WrittenDescription}_{\S112}(c, \mathrm{interp}, \mathrm{spec}(c))$ \Comment{specialization restated; no rule}
  \State $\mathrm{OA} \gets \textsc{OfficeAction}(c, \mathrm{nov}, \mathrm{obv}, \mathrm{wd})$
  \State $\mathrm{ver} \gets \textsc{VerifyRejections}(c, \mathrm{OA})$;\; $\mathrm{OA.grounds} \gets \mathrm{ver.verified}$ \Comment{verified grounds replace the draft}
  \State $\mathrm{fd} \gets \textsc{FinalDecision}(c, \mathrm{nov}, \mathrm{obv}, \mathrm{OA})$ \Comment{rule: strictness $\geq 0.7$ resolves uncertain grounds to reject; $\leq 0.4$ to grant or amend}
  \State $T \gets (\mathrm{interp}, \mathrm{map}, \mathrm{nov}, \mathrm{obv}, \mathrm{wd}, \mathrm{OA}, \mathrm{ver}, \mathrm{fd})$
  \State \Return $(\mathrm{fd.decision}, T)$ \Comment{decision $\in \{$grant, amend, reject$\}$}
\EndProcedure
\end{algorithmic}
\end{algorithm*}

\subsection{Pipeline Stages and MPEP Correspondence}

Each of the eight pipeline stages maps to a specific MPEP section and the statutory or case-law basis it implements. Table~\ref{tab:pipeline} lists the correspondence; stages 6--7 are simulation-specific quality controls without a direct one-to-one MPEP analogue.

\begin{table*}
\centering
\caption{Pipeline stages and MPEP correspondence.}
\label{tab:pipeline}
\footnotesize
\begin{tabular}{@{}cp{2.5cm}p{1.5cm}p{4cm}p{5cm}@{}}
\toprule
\textbf{\#} & \textbf{Pipeline step} & \textbf{MPEP} & \textbf{Legal basis} & \textbf{Notes} \\
\midrule
1 & Claim Interpretation & \S2111 & BRI doctrine & All examination begins with claim construction. BRI is the standard for examination (distinct from Phillips claim construction used in litigation). \\
2 & Prior Art Mapping & \S904 & Prior art search, element-by-element comparison & In our simulation, prior art is provided from PTAB case records, a limitation noted in Section~\ref{sec:limitations}. \\
3 & \S102 Novelty Assessment & \S2131 & 35 USC \S102: Anticipation & ``All elements'' test requires one-to-one correspondence. \\
4 & \S103 Obviousness Analysis & \S2141 & 35 USC \S103; Graham (1966); KSR (2007) & Four-factor analysis. \textbf{Stage where examiner discretion is greatest}, as the standard for ``motivation to combine'' is inherently subjective \citep{mandel2007patently}. \\
5 & \S112 Written Description & \S2161--2164 & 35 USC \S112(a),(b) & Relatively objective compared to \S103. \\
6 & Office Action Draft & \S707 & Standard format & Compiles all rejection grounds with specific claim citations and prior art references. \\
7 & Rejection Verification & N/A (sim.-specific) & Analogous to SPE review and OPQA quality checks & Simulation-specific quality filter for LLM hallucination mitigation. Differentiated standards: strict for \S102, permissive for \S103, strict for \S112. \\
8 & Final Decision & \S1302.14 & Examiner's final determination & Synthesizes all analyses; includes a weighing step informed by strictness. \\
\bottomrule
\end{tabular}
\end{table*}

\begin{table*}
\centering
\caption{Deviations from real USPTO examination practice.}
\label{tab:deviations}
\footnotesize
\begin{tabular}{@{}p{2.5cm}p{3.5cm}p{3.5cm}p{4cm}@{}}
\toprule
\textbf{Aspect} & \textbf{Real USPTO} & \textbf{Our simulation} & \textbf{Justification} \\
\midrule
Prior art search & Examiner searches EAST/WEST databases & Prior art provided from PTAB records & Standard in PTAB-based evaluation (PILOT-Bench follows same protocol) \\
Applicant interaction & Multiple rounds of OA $\rightarrow$ response $\rightarrow$ amendment & Single-pass examination & PTAB cases represent final-stage decisions \\
Examiner workload & 19.5 hours per disposal (average) & No time constraint & Time pressure modeled indirectly through seniority parameter \\
Institutional context & Office culture, SPE guidance, examiner meetings & Absent & Limitation noted in Section~\ref{sec:limitations} \\
Verification stage & No formal self-check (but SPE review exists) & Explicit verification stage (Stage 7) & Simulation-specific quality filter for LLM hallucination mitigation \\
Interview practice & Examiner-applicant interviews are common & Not modeled & Limitation; interviews can significantly affect outcomes \\
\bottomrule
\end{tabular}
\end{table*}

\subsection{Examiner Dimensions --- Empirical and Legal Grounding}

Each of the five examiner dimensions was selected because it has been identified as a statistically significant predictor of examination outcomes in peer-reviewed work; each is grounded in a specific legal standard and is parameterized numerically at a specific point in the pipeline. \textit{Strictness} has no explicit MPEP standard and is the source of the ``patent lottery'' \citep{lemley2012examiner, frakes2015patent}; it modulates the confidence threshold at the final decision stage, with values $\geq 0.7$ activating a ``public protection'' framing and $\leq 0.4$ an ``innovation support'' framing. \textit{Claim breadth} reflects the BRI standard's ``broadest reasonable interpretation'' (MPEP \S2111), whose word ``reasonable'' introduces interpretive discretion; it enters the claim interpretation stage as a numeric parameter, where a broader reading widens the prior-art attack surface at \S\S102--103. \textit{Combination threshold} operationalizes the expansion of permissible obviousness grounds under \textit{KSR v.~Teleflex} (2007) \citep{mandel2007patently}; it directly modulates \S103, with low values accepting common-sense combinations and high values requiring explicit teaching. \textit{Seniority} reflects the USPTO GS-grade system, where promotion reduces per-case time and raises grant rates \citep{frakes2017time}; its value is carried in the persona block, annotated with the pattern-recognition and time-pressure profile of senior examiners \citep{klein1998sources}. \textit{Technology specialization} reflects Art-Unit assignment by domain, where specialists are more consistent and non-specialists show grant bias \citep{righi2019examiner}; it is set numerically at the \S112 stage, which calibrates the PHOSITA standard.

\subsection{Analytical Skills --- Legal Framework Mapping}

\begin{table}[!ht]
\centering
\caption{Cohort stratification across mutually exclusive PTAB outcome categories. ``With new grounds'' aggregates the three new-ground-of-rejection sub-categories (Affirmed/Reversed/Affirmed-in-Part with new grounds). 54/100 (54\%) are boundary cases where PTAB reversed or partially reversed the examiner's decision.}
\label{tab:cohort}
\footnotesize
\setlength{\tabcolsep}{4pt}
\begin{tabular}{@{}lccc@{}}
\toprule
\textbf{PTAB Outcome} & \textbf{Pool} & \textbf{Selected} & \textbf{Enrichment} \\
\midrule
Affirmed & 4,208 (54.7\%) & 45 (45\%) & Under-sampled \\
Reversed & 2,257 (29.3\%) & 35 (35\%) & Enriched \\
Affirmed-in-Part & 835 (10.8\%) & 15 (15\%) & Enriched \\
With new grounds & 398 (5.2\%) & 5 (5\%) & Proportional \\
\midrule
Total & 7,698 (100\%) & 100 & --- \\
\bottomrule
\end{tabular}
\end{table}

All examiners share the same four analytical skills, each tied to its governing legal test and each receiving specific dimensions as numeric inputs. \textit{\S102 novelty} applies the ``all elements'' test (MPEP \S2131; \textit{In re Robertson}), asking whether a single reference discloses every claim element; the claims as interpreted under the examiner's breadth parameter fix the element scope, and the strictness value is passed to this stage. \textit{\S103 obviousness} applies the Graham four-factor analysis and the seven KSR rationales for motivation to combine (MPEP \S2141; \textit{Graham v.~John Deere}; \textit{KSR v.~Teleflex}); the combination threshold is its rule-based modulator, and claim breadth enters through the interpreted claims that define the differences at Factor 2. \textit{\S101 eligibility} enters through the persona block, which instructs every examiner to flag claims that may recite an abstract idea, and through the verification stage, which keeps such a flag when its \textit{Alice}/\textit{Mayo} reasoning (MPEP \S2106) is sound. The flag was raised in 2 of 1{,}400 decisions, only once on the 29 cases whose PTAB decision involved \S101 (Limitations). \textit{\S112 written description} covers written description, enablement, and definiteness (MPEP \S2161--2164; \textit{Ariad v.~Eli Lilly}); tech specialization is the parameter passed to this stage, which calibrates the PHOSITA standard, and this skill shows the most convergent results across profiles, consistent with its relatively objective nature.

\subsection{Deviations from Real Examination}

For transparency, we document where our simulation departs from actual USPTO practice; Table~\ref{tab:deviations} lists each deviation with its justification.

\section{Cohort Stratification Details}
\label{app:cohort}

\textit{Source pool.} Starting from the 18,049 PTAB decisions in PILOT-Bench, we retrieve and parse the USPTO publications for each applicant and prior-art patent number. This yields 15,531 cases with full applicant specifications; of these, 7,698 also have at least one prior-art patent with retrievable specifications and form our source pool (excluding 10 development cases used for pipeline calibration). The full-text corpus is released on Hugging Face (Appendix~\ref{app:fullresults}).

\textit{Stratified selection.} We draw 100 cases enriched for boundary cases to maximize statistical power for disagreement analysis (Table~\ref{tab:cohort}). Outcome categories below are mutually exclusive, following the PTAB \texttt{subdecisionTypeCoarse} taxonomy:

\textit{Technology center distribution} (Table~\ref{tab:tc}):

\begin{table}[!ht]
\centering
\caption{Technology center distribution of the 100-case cohort.}
\label{tab:tc}
\small
\begin{tabular}{@{}clr@{}}
\toprule
\textbf{TC} & \textbf{Domain} & \textbf{Cases} \\
\midrule
2400 & Computer Networks, Multiplex Comm. & 22 \\
3600 & Transportation, E-Commerce, Agri. & 21 \\
2800 & Semiconductors, Electrical Systems & 15 \\
1700 & Chemical and Materials Engineering & 14 \\
3700 & Mechanical Engineering, Manufacturing & 11 \\
2600 & Communications & 8 \\
2100 & Computer Architecture, Software & 8 \\
1600 & Biotechnology, Organic Chemistry & 1 \\
\bottomrule
\end{tabular}
\end{table}

\textit{Issue distribution:} \S103 (obviousness): 90 cases. \S102 (novelty): 30. \S101 (eligibility): 29. \S112 (written description): 29. Multi-issue: 58 cases (58\%).

\textit{Selection bias check.} Because we overrepresent boundary cases, aggregate disagreement \textit{rates} in our cohort should not be extrapolated to the full PTAB population. However, findings about disagreement \textit{structure}, namely which dimensions drive disagreement, how disagreement cascades through the pipeline, and whether coverage holds, are properties of the simulation design and should generalize beyond this specific cohort composition.

\section{Baseline Comparison --- Vanilla and Naive Persona}
\label{app:baseline}

To verify that the observed examiner diversity arises from our orthogonal design rather than LLM stochasticity or simple persona prompting, we run two baseline experiments on the same 10 pilot cases (140 decisions each) using the identical examination pipeline and LLM backbone (Qwen3-235B-A22B, temperature 0.4, seed 42). These baselines serve as diagnostic comparisons, designed to test whether the structured design is necessary for the observed diversity patterns, not as matched-scale alternative experiments. Ten cases are sufficient for this diagnostic purpose because the failure modes we test (label compliance, content sensitivity, structural attributability) are observable at the individual-case level.

\subsection{Baseline Designs}

\textit{Vanilla baseline.} All 14 examiners use identical default profiles (all five dimensions = 0.50, noise $\sigma = 0.05$), with no archetype differentiation. The resulting strictness range is 0.426--0.599 (vs.\ 0.25--0.89 in the main experiment). Any observed disagreement reflects only LLM sampling noise and minor parameter perturbation.

\begin{table}[h]
\centering
\caption{Per-case results for vanilla baseline (14 examiners, all dimensions $\approx 0.50$). Aggregate: 111 reject (79.3\%), 2 grant (1.4\%), 27 amend (19.3\%). All 10 cases show disagreement.}
\label{tab:vanilla}
\small
\begin{tabular}{@{}lccccc@{}}
\toprule
\textbf{Case} & \textbf{GT} & \textbf{R} & \textbf{G} & \textbf{A} & \textbf{Disagree} \\
\midrule
2017000821 & Aff. & 6  & 0 & 8 & Yes \\
2017006232 & Rev. & 12 & 0 & 2 & Yes \\
2018001417 & Aff. & 13 & 0 & 1 & Yes \\
2018003455 & Aff. & 11 & 0 & 3 & Yes \\
2018006721 & Aff. & 11 & 0 & 3 & Yes \\
2019006408 & Rev. & 13 & 0 & 1 & Yes \\
2020001686 & Aff. & 12 & 0 & 2 & Yes \\
2020006223 & Rev. & 12 & 1 & 1 & Yes \\
2021003412 & Rev. & 13 & 0 & 1 & Yes \\
2021004207 & Rev. & 8  & 1 & 5 & Yes \\
\bottomrule
\end{tabular}
\end{table}

\textit{Naive persona baseline.} 14 examiners are split into two groups: 7 receive a strict persona label (``You are a very strict patent examiner who tends to reject applications'') and 7 receive a lenient persona label (``You are a very lenient patent examiner who tends to grant applications''). All quantitative dimension values remain at 0.50. This design mirrors the persona-prompting approach common in LLM simulation literature (Section~\ref{sec:related-llmgen}): diversity is introduced through natural language instruction rather than structured skill parameters.

Both baselines use the same 8-stage examination pipeline as the main experiment: claim interpretation $\rightarrow$ prior art mapping $\rightarrow$ novelty assessment $\rightarrow$ obviousness analysis $\rightarrow$ written description check $\rightarrow$ office action $\rightarrow$ rejection verification $\rightarrow$ final decision. The only difference is how examiner profiles are configured.

\subsection{Vanilla Baseline Results}

\textit{Key finding: diversity exists but is unstructured.} The vanilla baseline (Table~\ref{tab:vanilla}) produces a G+A spread of 40.0 pp across examiners (range: 0.0\%--40.0\%): less than half of the main experiment's 92.5 pp spread. More critically, splitting examiners at the median strictness yields an effect of only $-$4.3 pp, compared to $-$40.0 pp for the main experiment under the same split. The direction of individual examiner decisions is not predictable from profile parameters: an examiner with strictness 0.426 shows 40.0\% G+A rate, while an examiner with strictness 0.474 shows 30.0\%, so the noise-level parameter differences do not systematically explain who grants and who rejects.

This confirms that the LLM backbone itself introduces stochastic variation, but this variation is (a) smaller in magnitude, (b) reject-skewed (79.3\% vs.\ the main experiment's more balanced distribution), and (c) not attributable to examiner dimensions: precisely the pattern expected when no structured design is imposed.

\begin{table}[h]
\centering
\caption{Per-case results for persona baseline (7 strict-labeled + 7 lenient-labeled examiners). Aggregate across 140 decisions: strict-labeled examiners reject 85.7\% (60/70), grant 0\% (0/70), amend 14.3\% (10/70); lenient-labeled examiners reject 11.4\% (8/70), grant 42.9\% (30/70), amend 45.7\% (32/70). G+A spread: 100.0 pp. Label effect: 74.3 pp.}
\label{tab:persona}
\small
\begin{tabular}{@{}lccc@{}}
\toprule
\textbf{Case} & \textbf{GT} & \textbf{Strict R/G/A} & \textbf{Lenient R/G/A} \\
\midrule
2017000821 & Aff. & 3/0/4 & 0/7/0 \\
2017006232 & Rev. & 6/0/1 & 2/1/4 \\
2018001417 & Aff. & 7/0/0 & 0/3/4 \\
2018003455 & Aff. & 7/0/0 & 4/0/3 \\
2018006721 & Aff. & 7/0/0 & 0/4/3 \\
2019006408 & Rev. & 7/0/0 & 1/2/4 \\
2020001686 & Aff. & 7/0/0 & 1/1/5 \\
2020006223 & Rev. & 7/0/0 & 0/4/3 \\
2021003412 & Rev. & 6/0/1 & 0/2/5 \\
2021004207 & Rev. & 3/0/4 & 0/6/1 \\
\bottomrule
\end{tabular}
\end{table}

\subsection{Naive Persona Baseline Results}

\textit{Key finding: persona labels produce diversity, but the diversity is label-driven rather than content-driven.} Three observations support this (Table~\ref{tab:persona}):

\begin{enumerate}
\item \textit{Strict examiners never grant.} Across all 10 cases and 70 strict-labeled decisions, the grant count is exactly zero. This includes 5 Reversed cases where PTAB determined the application should have been granted; yet no strict-labeled examiner ever reaches this conclusion. The label overrides content analysis.
\item \textit{Lenient examiners grant Affirmed cases.} In the 5 Affirmed cases (where PTAB confirmed the rejection), lenient-labeled examiners produce 15 grants and 15 amends versus only 5 rejects. For case 2017000821, all 7 lenient examiners grant; yet the main experiment and vanilla baseline both produce zero grants for this case.
\item \textit{The label effect (74.3 pp) exceeds the main experiment's strictness effect ($-$33.6 pp).} This is not evidence that persona labels are more powerful than structured parameters; it is evidence that labels bypass content analysis. In the main experiment, even the most extreme archetype (C1 veteran strict, 7.0\% G+A) still grants some cases where the evidence warrants it. Persona strict examiners grant none.
\end{enumerate}

\begin{table}[h]
\centering
\caption{Single-case comparison across all conditions (case 2017000821, GT=Affirmed). Both main and vanilla produce \textbf{zero grants}; both persona conditions produce \textbf{6--7 grants}, all from lenient-labeled examiners.}
\label{tab:fourway}
\small
\begin{tabular}{@{}lccc@{}}
\toprule
\textbf{Condition} & \textbf{R} & \textbf{A} & \textbf{G} \\
\midrule
Main (orth., Qwen3) & 7 & 7 & 0 \\
Vanilla (all 0.50, Qwen3) & 6 & 8 & 0 \\
Persona (Qwen3) & 3 & 4 & 7 \\
Persona (GPT-4.1) & 2 & 6 & 6 \\
\bottomrule
\end{tabular}
\end{table}

The contrast with the main experiment is instructive. Our strict gatekeeper (strictness=0.85) rejects most cases but grants when the evidence is strong; our lenient advocate (strictness=0.30) grants most cases but rejects when the evidence is overwhelming. Both respond to case content through their skill parameters. Persona-labeled examiners respond primarily to the label instruction, producing diversity that is wider in spread but shallower in structure: the disagreement cannot be traced to specific analytical differences, only to the binary strict/lenient assignment.

\subsection{Case Study: Why Persona Grants Are Content-Blind (Case 2017000821)}

To make the persona failure concrete, we examine case 2017000821 in detail. This patent application, ``Insulated Electric Motor and Method of Forming,'' claims a method of filling gaps between stator winding pairs using insulating felt or aramid-based adhesive materials. The single prior art reference (US-2012165198-A1) already discloses stator slot gaps with thermally conductive fill material. The core \S103 question is whether substituting insulating felt for thermal material is obvious to a person of ordinary skill in the art. PTAB affirmed the rejection, finding the material substitution obvious.

Our structured examiners respond to the case evidence: the strict gatekeeper rejects because ``the gap-filling structure is well-known and the material substitution is obvious''; the lenient advocate recommends amend because ``the dual-material layered filling is not explicitly taught, so narrowing the claims could distinguish over prior art.'' Both positions are legally grounded in the case evidence. The persona lenient examiners, by contrast, grant the application outright: a position that no content-analyzing examiner (structured or vanilla) reaches, because the prior art coverage is too close.

To test whether this failure is model-specific, we ran the same persona configuration on the same case using GPT-4.1 (OpenAI, April 2025); results across all four conditions are in Table~\ref{tab:fourway}.

\textit{The persona failure mode differs by model but the limitation is shared.} Qwen3's strict-labeled examiners interpret ``strict'' as an instruction to reject or scrutinize (3/7 reject, 4/7 amend), while GPT-4.1's strict-labeled examiners interpret ``strict'' as an instruction to examine carefully (6/7 amend). The natural language label ``strict'' is semantically ambiguous (does it mean ``reject more,'' ``scrutinize more carefully,'' or ``apply higher standards''?), and different models resolve this ambiguity differently. Meanwhile, both models' lenient-labeled examiners converge on grant regardless of case content, suggesting that ``lenient'' functions as a stronger, less ambiguous instruction than ``strict.''

This asymmetry (where one pole of a persona label is interpretable and the other is not) is precisely the kind of uncontrollable behavior that makes persona-based diversity unsuitable for structured judgment space generation. Our skill-parameter approach avoids this entirely: strictness=0.85 operates numerically on the final decision synthesis stage, leaving no room for semantic reinterpretation.

This cross-model evidence supports two claims: (1) persona limitations are not Qwen3-specific but reflect a fundamental constraint of natural language label-based diversity; and (2) our single-LLM design choice is partially mitigated by the fact that even switching to a different frontier model does not resolve the persona problem: the structured design, not the model, is the key factor.

\subsection{Three-Way Comparison}

\begin{table*}[t]
\centering
\caption{Diagnostic comparison of the main orthogonal design (100 cases) with the vanilla and persona baselines (10 pilot cases each). The effect row is the G+A rate of the strict half of the population minus that of the lenient half, split at the median strictness for Main and Vanilla and by label for Persona.}
\label{tab:threeway}
\small
\begin{tabular}{@{}lp{3cm}p{3.8cm}p{4.5cm}@{}}
\toprule
\textbf{Metric} & \textbf{Main (orthogonal)} & \textbf{Vanilla (all 0.50)} & \textbf{Persona (strict/lenient)} \\
\midrule
G+A spread (pp) & 92.5 & 40.0 & 100.0 \\
Strictness/label effect (pp) & $-$40.0 & $-$4.3 & $-$74.3 \\
Disagreement rate & 99/100 & 10/10 & 10/10 \\
Profile-attributable? & Yes (37/37, $p<0.0001$) & No (direction inconsistent) & No (label-driven, not content-driven) \\
Decision distribution & Balanced & Reject-skewed (79.3\%) & Polarized (strict=86\% R, lenient=89\% G+A) \\
Strict ever grants? & Yes (content-responsive) & N/A & \textbf{No (0/70)} \\
Content-sensitive? & Yes & Partially & No (Affirmed cases get grants) \\
\bottomrule
\end{tabular}
\end{table*}

The comparison (Table~\ref{tab:threeway}) reveals three distinct diversity regimes. The \textit{vanilla baseline} produces stochastic diversity: disagreement exists but is unstructured, reject-skewed, and not attributable to examiner parameters. The \textit{persona baseline} produces label-driven diversity: disagreement is wide (100 pp spread) but follows the binary strict/lenient assignment rather than case content: strict examiners never grant regardless of evidence, and lenient examiners grant even Affirmed cases. The \textit{orthogonal design} produces structured, content-responsive diversity: disagreement traces to specific examiner dimensions, respects case evidence (even extreme archetypes respond to content), and supports process-level explanation through the cascade mechanism. In short, \textit{stochastic diversity is noise; label-driven diversity is artifact; only structured diversity enables judgment space generation.}

\begin{table}[h]
\centering
\caption{Effect of institutional oversight on per-archetype favorable rate (grant + amend) across 50 shared cases. Moderate archetypes show the largest reductions, three of four at $p < 0.01$. Among the extremes, O1 reaches significance while C1 and C2 do not.}
\label{tab:oversight-full}
\small
\setlength{\tabcolsep}{3pt}
\begin{tabular}{@{}llcccc@{}}
\toprule
\textbf{ID} & \textbf{Archetype} & \textbf{Base.} & \textbf{Treat.} & \textbf{$\Delta$} & \textbf{$p$} \\
\midrule
\multicolumn{6}{@{}l}{\textit{Moderate}} \\
O2 & broad\_interpreter  & 0.530 & 0.320 & $-$0.210 & 0.0004 \\
O3 & easy\_combiner      & 0.530 & 0.400 & $-$0.130 & 0.0590 \\
O4 & senior\_examiner    & 0.540 & 0.360 & $-$0.180 & 0.0012 \\
O5 & domain\_specialist  & 0.480 & 0.280 & $-$0.200 & 0.0030 \\
\midrule
\multicolumn{6}{@{}l}{\textit{Extremes}} \\
O1 & strict\_gatekeeper  & 0.130 & 0.060 & $-$0.070 & 0.0155 \\
C1 & veteran\_strict     & 0.090 & 0.040 & $-$0.050 & 0.2742 \\
C2 & lenient\_advocate   & 1.000 & 0.970 & $-$0.030 & 0.2497 \\
\bottomrule
\end{tabular}
\end{table}

\section{Institutional Oversight Experiment --- Per-Archetype Details}
\label{app:oversight-full}

Table~\ref{tab:oversight-full} reports per-archetype baseline (control) and treatment favorable rates (grant + amend) for the institutional oversight experiment described in Section~\ref{sec:oversight}. The treatment condition adds an institutional oversight prompt to the examination pipeline; all other parameters are held fixed. Statistical significance is assessed via paired permutation test (1{,}000{,}000 sign-flips, seed = 42) over the 50 shared cases. The test unit is the per-case favorable fraction of each archetype's two instances, paired across conditions.

\begin{table}[h]
\centering
\caption{Population-level simulation of professional judgment across domains.}
\label{tab:related}
\small
\resizebox{\columnwidth}{!}{%
\begin{tabular}{@{}p{2.2cm}lp{1.6cm}p{1.5cm}@{}}
\toprule
\textbf{Study} & \textbf{Venue} & \textbf{Diversity method} & \textbf{Scale} \\
\midrule
AgentReview \citep{jin2024agentreview} & EMNLP 2024 & Three reviewer dims & 3,797 papers \\
AgentCourt \citep{chen2025agentcourt} & ACL 2025 Find. & Adversarial evolution & Court cases \\
Law in Silico \citep{wang2025law} & ACL 2026 Find. & Statistical sampling & Society-level \\
PatientSim \citep{kyung2025patientsim} & NeurIPS 2025 D\&B & Persona-driven & Doctor-patient \\
MoP \citep{bui2025mop} & ACL 2025 Find. & Probabilistic mixture & Survey resp. \\
Homo Silicus \citep{horton2024homo} & EC 2024 & Conditional prompting & Survey resp. \\
\bottomrule
\end{tabular}%
}
\end{table}

\textit{Reading the table.} The four moderate archetypes (O2, O3, O4, O5) lose 13--21 pp of favorable rate under oversight, with three reaching $p < 0.01$. The two strict extremes (O1, C1) start near the floor (9--13\%), so their absolute drops are small. O1's 7 pp drop is nonetheless significant ($p = 0.016$), while C1's is not ($p = 0.27$). Relative to the room each archetype had to fall, the strict extremes lose 54\% (O1) and 56\% (C1) of their baseline favorable rate, more than the moderate archetypes (25--42\%). The lenient extreme (C2) starts at the ceiling (100\%) and drops 3 pp ($p = 0.25$), 3\% of its available range.

\section{Additional Related Work}
\label{app:related}

This appendix expands on the related work in Section~\ref{sec:related}. We first situate our design against prior LLM-based simulations of professional judgment, then detail the cognitive-science basis for modeling examiner diversity through skill parameters rather than persona labels.

\subsection{LLM Simulation Comparison}

A common feature of the approaches in Table~\ref{tab:related} is that agent diversity is modeled through persona descriptions or role assignments, and the primary goal is measuring the \textit{extent} of variation or bias. This framing has been critiqued on several grounds. \citet{lin2025six} identifies six fallacies of LLM-based human substitution, including the ``essentialization'' of identity through simple trait labels. \citet{agnew2024illusion} argue that LLM-simulated populations should complement, not replace, real human participants. Recent work in social simulation identifies a ``diversity collapse'' phenomenon, where persona-prompted agents converge to similar outputs despite different trait labels \citep{wang2026parametric}. Our work addresses these critiques through skill-parameter-based diversity, orthogonal experimental design, and a judgment-space-generation framing rather than human substitution (see Section~\ref{sec:related-llmgen}).

\subsection{Cognitive Science of Expert Disagreement (Extended)}

Our approach to modeling examiner diversity draws on three streams of research in cognitive science. The Dreyfus model of skill acquisition \citep{dreyfus1986mind} establishes that expertise develops through qualitatively distinct stages, from rule-following to pattern-based intuition: implying that differences among experts reflect differences in \textit{what patterns they recognize}, not personality traits. Klein's Recognition-Primed Decision model \citep{klein1998sources} extends this by showing, through naturalistic studies of firefighters, nurses, and military commanders, that experienced professionals make different decisions because they have constructed different mental models from different accumulated experiences. \citet{kahneman2009conditions} identify two conditions for reliable expert intuition: a regular environment and adequate opportunity for learning through feedback. Patent examination, with its delayed and incomplete feedback loop, only partially satisfies these conditions, providing a theoretical basis for expecting persistent, stable inter-examiner disagreement.

Together, these findings motivate our core modeling choice: rather than varying examiner ``personality'' through trait labels, we configure examiners with the same legal frameworks but different application thresholds, reflecting how real examiners all follow the MPEP but exercise judgment differently.

\end{document}